\documentclass{article}
\usepackage{graphicx}
\usepackage{soul}
\usepackage{xcolor}     % 颜色包（高亮核心）
\usepackage{booktabs}
\usepackage{multirow}
\usepackage{graphicx}
\usepackage{subcaption}
\usepackage{listings}
\usepackage{verbatim}
\usepackage{tcolorbox}
\usepackage{hyperref}
\usepackage{enumitem}
\usepackage{tabularx}
\usepackage{wrapfig}
\usepackage{pgfplots}
\pgfplotsset{compat=1.18}
\usepackage{amssymb}
\usepackage{amsmath}
\usepackage{caption}
\usepackage[dvipsnames]{xcolor}
\usepackage{CJKutf8}
\definecolor{g_sum}{HTML}{3479E5}

\tcbuselibrary{breakable}
\usepackage[preprint]{corl_2026} % Uncomment for pre-prints (e.g., arxiv); This is like ``final'', but will remove the CORL footnote.

\title{Cortex: A Bidirectionally Aligned Embodied Agent Framework for Long-horizon Manipulation}

\author{
  \textbf{Jiaqi Peng$^{*\,1,2}$ \quad Xiqian Yu$^{*\,2}$ \quad Delin Feng$^{*\,2}$ \quad Yuqiang Yang$^{2}$ \quad Wenzhe Cai$^{2}$} \\
  \textbf{Jing Xiong$^{2,3}$ \quad Ganlin Yang$^{2,4}$ \quad Jinliang Zheng$^{1,2}$ \quad Jiafei Cao$^{2}$ \quad Xueyuan Wei$^{2}$} \\
  \textbf{Jiangmiao Pang$^{2}$ \quad Yuan Shen$^{\dagger\,1}$ \quad Tai Wang$^{\dagger\,2}$} \\[8pt]
  $^{1}$Tsinghua University \quad $^{2}$Shanghai AI Laboratory \quad $^{3}$Peking University \quad $^{4}$USTC\\
  {\tt \href{https://steinate.github.io/cortex.github.io}{\textcolor{g_sum}{https://steinate.github.io/cortex.github.io}}}\\
  \textcolor{gray!70}{$^*$Equal contribution. $^\dagger$Corresponding author.}
  \vspace{-30pt}
}

\begin{document}
\begin{CJK*}{UTF8}{gbsn}
\maketitle
% \blfootnote{$^*$Equal contribution. $^\dagger$Corresponding author.}

\begin{abstract}
While recent Vision-Language-Action (VLA) models show promise toward generalist manipulation policies, they struggle with long-horizon tasks due to their Markovian nature—relying solely on current observations. Hierarchical dual-system methods address this but suffer from a gap between high-level planning semantics and low-level execution kinematics. We introduce Cortex, a bidirectionally aligned embodied agent framework with a customized planning interface that conveys \emph{executable} and \emph{tractable} subtask plans from high-level VLM to low-level VLA. Specifically, we standardize manipulation subtasks into 32 canonical skill primitives and inject tractability principles, such as representative object attributes and improved trajectory reachability, into the data generation pipeline.
This enables automatic annotation of over 4k hours of open-source video data and generation of 30 hours of simulation data. We further devise an event-balanced sampling strategy to construct training data for fine-tuning the framework to better handle planning ambiguity during subtask transitions, enhanced by carefully designed harness engineering from task contexts to skill constraints during inference. Both open-loop VLM and closed-loop system evaluations demonstrate Cortex's efficacy, \emph{e.g.}, it outperforms monolithic baselines by 3.1\% on Libero-long and 4.1\% on RoboTwin. Notably, Cortex's generalist VLM enables zero-shot completion of unseen real-world long-horizon tasks, such as multi-stage chemistry experiments, by simply combining with a fine-tuned VLA—a capability infeasible through VLA fine-tuning alone.

\end{abstract}


\vspace{-5mm} 
\begin{figure}[htbp]
\centering
\includegraphics[width=0.94\textwidth]{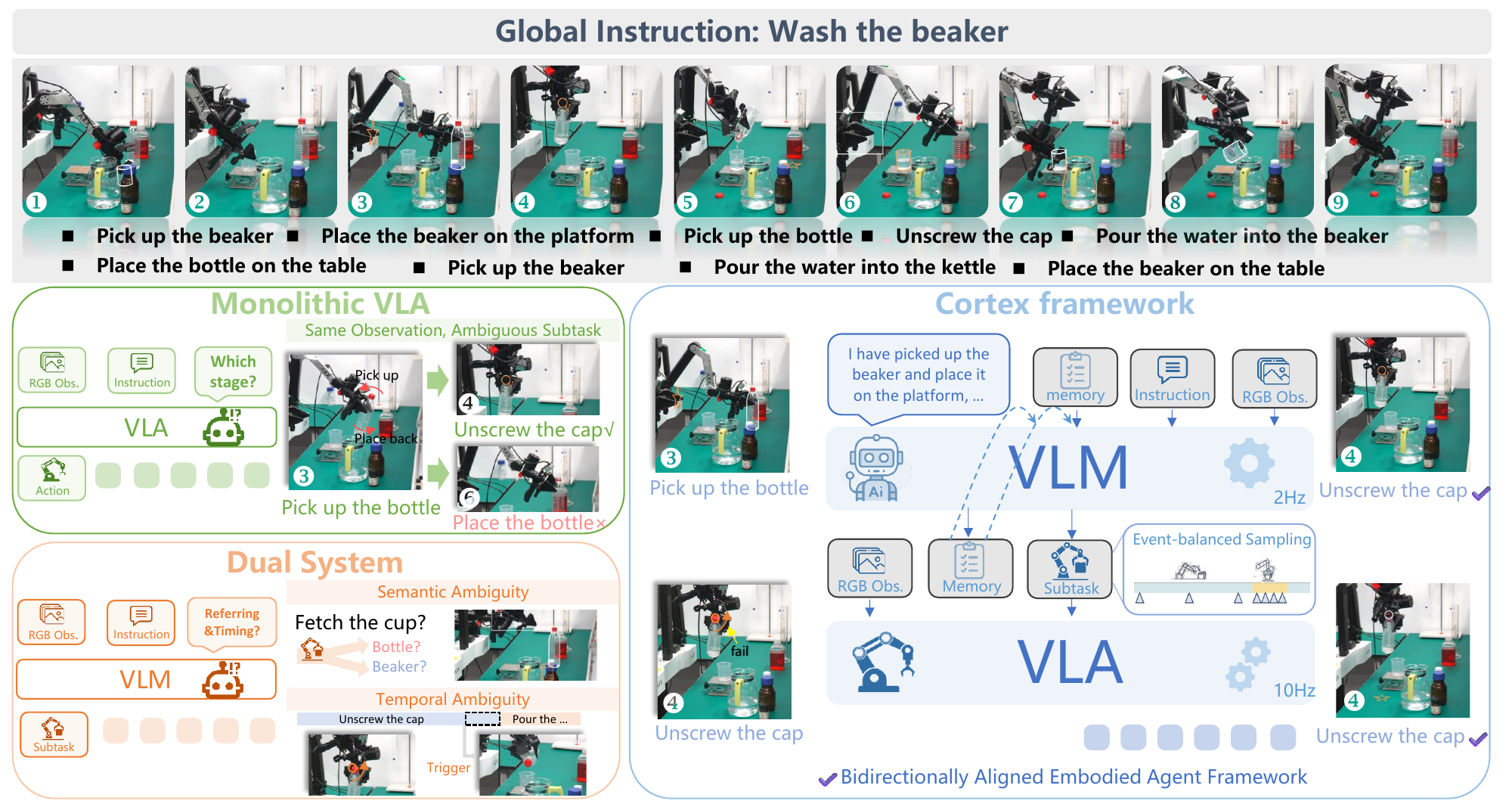}
\caption{Compared to previous works, Cortex is an embodied agent framework aligning the high-level VLM and low-level VLA on the subtask executability and tractability, enabling synergistic planning and execution for long-horizon manipulation.}
\vspace{-5mm} 
\end{figure}

\section{Introduction}

Vision-Language-Action (VLA) models have fundamentally reshaped embodied AI by directly mapping multi-modal inputs to continuous motor control, achieving remarkable zero-shot generalization in short-horizon tasks \cite{hurst2024gpt,chen2024internvl,bjorck2025gr00t,kim2024openvla}. However, as tasks scale in temporal complexity, monolithic VLAs face a critical bottleneck: Markovian short-sightedness. Operating purely reactively without continuous progress verification or spatial-temporal memory, these models struggle to differentiate actual task progress from instantaneous visual observations \cite{torne2026mem,liu2026long}. Consequently, when executing monolithic long-horizon instructions, they frequently lose track of intermediate states and blindly repeat actions, inevitably leading to compounding execution errors.

Recent efforts to mitigate this via visual frame buffering \cite{shi2025memoryvla,li2025cronusvla} are constrained by limited context windows and struggle to form the semantic memory required for logical planning. Alternatively, hierarchical dual-system paradigms decouple cognitive planning from reactive execution. However, they fundamentally suffer from a lack of bidirectional alignment, \emph{i.e.}, the high-level VLM should consider the capabilities of low-level VLA, while the VLA should be robust and adaptive to the VLM's outputs. Early dual-system planners \cite{ahn2022can,huang2022inner} act as disembodied observers: The output plans lack explicit embodied constraints, and thus serve as kinematically ungrounded instructions for downstream executors, leading to a severe semantic-kinematic domain gap. Recent works~\cite{torne2026mem,sridhar2025memer,chen2026rmbench} equip the VLAs with the memorization capability but still fall short in the planning alignment aspect.

To tackle these challenges, we propose \textbf{Cortex}, a \underline{c}ognitive \underline{or}chestra\underline{t}or aligned with \underline{ex}ecution.
It is achieved by a bidirectionally aligned embodied agent framework featuring a customized planning interface that conveys \emph{executable} and \emph{tractable} information from the high-level VLM to the low-level VLA.
Specifically, we first define an interface subtask formulation and construct the metadata accordingly.
For \textbf{executability}, we standardize manipulation subtasks into 32 canonical skill primitives, enabling the automatic annotation of over 4k hours of open-source video data. For \textbf{tractability}, we inject physical principles—such as representative object attributes and improved trajectory reachability—directly into the data generation pipeline, yielding over 30 hours of high-quality simulation data.
Built upon these metadata, we further devise an event-balanced sampling strategy to construct subtask execution and transition samples for fine-tuning the framework to better handle the planning ambiguity during different phases.
During inference, we augment the agent framework with carefully designed harness engineering, involving task contexts and skill constraints to ensure the alignment between planners and executors.

Both open-loop VLM and closed-loop dual system evaluations validate the efficacy of Cortex. In the open-loop VLM evaluation, our Cortex's VLM outperforms state-of-the-art generalist models including GPT-5~\cite{gpt5systemcard2025} and Gemini~\cite{geminipro31preview2026}. In closed-loop simulation benchmarks, Cortex achieves state-of-the-art performance, reaching a 95.5\% success rate (+3.1\%) on Libero-long and 86.8\% (+4.1\%) on RoboTwin, decisively outperforming its underlying monolithic VLA baseline ($\pi_{0.5}$). More importantly, Cortex's generalist VLM enables the zero-shot completion of unseen real-world long-horizon tasks, such as complex multi-stage chemistry experiments, by simply combining with a fine-tuned VLA—a capability fundamentally infeasible through VLA fine-tuning alone. Compared to the zero success rate of end-to-end methods, Cortex achieves up to 65\% success rate on such tests, which is even comparable to the performance of humans with VLA policies.

\section{Related Work}
\label{sec:related_work}

\textbf{Vision-Language-Action Models.} 
Vision-Language-Action (VLA) models reframe continuous control as multimodal sequence modeling. Early milestones like RT-1 \citep{zitkovich2023rt} and PaLM-E \citep{driess2023palm} integrated sensory streams with language, while RT-2 \citep{zitkovich2023rt} co-fine-tuned web-scale VLMs directly on robotic data for semantic-to-motor translation. Recently, diverse open-weight frameworks (Octo \citep{team2024octo}, OpenVLA \citep{kim2024openvla}, InternVLA-M1 \citep{chen2025internvla}) and foundation models ($\pi_0$ \citep{intelligence2025pi05}, GR00T \citep{bjorck2025gr00t}) have achieved robust zero-shot and cross-embodiment generalization. However, despite excelling at short-horizon primitives, these monolithic end-to-end architectures remain fundamentally constrained by Markovian assumptions, severely limiting their performance on long-horizon tasks.

\textbf{Hierarchical Planning and Dual-System.} 
To address monolithic limitations, ``System-1/System-2'' paradigms decouple cognitive planning from reactive execution. Early methods like SayCan \citep{ahn2022can} and Code as Policies \citep{liang2023code} used explicit linguistic or code interfaces to orchestrate predefined skills. To bypass rigid textual boundaries, recent frameworks couple slow orchestrators with fast executors via continuous latent spaces for manipulation \citep{bu2024towards,zhang2024hirt} and navigation \citep{wei2025ground}. However, latent-coupled systems sacrifice interpretability and explicit progress tracking, while unconstrained language-coupled systems often fail to physically ground their plans. Crucially, both paradigms largely operate open-loop, lacking continuous visual success verification.

\textbf{Embodied Agents and Temporal Memory.}
Long-horizon autonomy demands persistent state representations. While classical Task and Motion Planning (TAMP) \cite{kaelbling2011hierarchical,garrett2021integrated} enables compositional reasoning, it struggles in partially observable, open-world environments. To maintain temporal consistency, recent embodied agents \cite{li2026roboclaw,xu2026roboagent,srivastava2014combined,fang2019scene} augment large models with memory mechanisms via textual histories, episodic retrieval, or visual buffering. Yet, existing approaches \cite{zhang2025memory} primarily treat memory as passive observation storage. They lack an action-oriented, physically grounded state abstraction necessary to resolve ambiguities during dynamic, multi-stage interactions.
%===============================================================================
\begin{figure}[!t]
\centering
\includegraphics[width=1.0\textwidth]{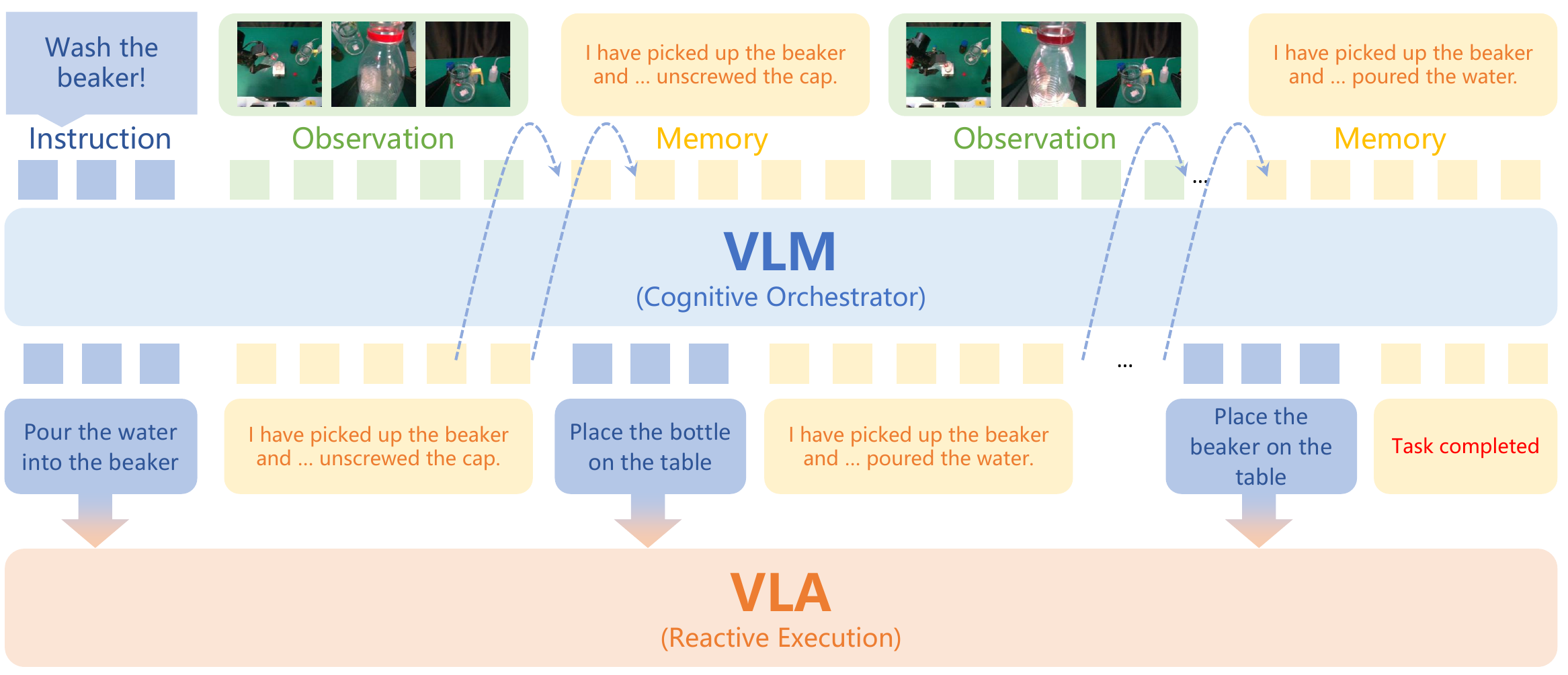}
% \caption{The Bidirectionally Aligned Subtask Interface. The cognitive orchestrator dynamically updates its semantic memory to continuously stream executable subtask primitives ($s_t$) to ground the reactive System-1 executor.}
\caption{Overview of the Cortex framework. The cognitive orchestrator dynamically updates semantic memory to continuously stream executable subtask to ground the reactive System-1 executor.}
% \caption{Overview of the Cortex framework. The system bidirectionally aligns a high-level Vision-Language Model (VLM) cognitive orchestrator with a low-level Vision-Language-Action (VLA) reactive executor.}
\label{fig:architecture}
\vspace{-5mm} 
\end{figure}

\section{Approach}
This section delineates the Cortex framework by first establishing a bidirectionally aligned subtask interface via scalable metadata construction, then introducing an event-balanced episode sampling strategy for ambiguity-resolving training, and finally deploying an asynchronous inference pipeline with harness engineering for closed-loop execution.

\subsection{Framework Overview}
As Fig.~\ref{fig:architecture} shows, the Cortex framework consists of a VLM and VLA. The VLM takes the instruction and image observations as input and performs subtask planning with textual memory records. The VLA takes the subtask as input for reactive execution. The key problem here is to define a customized subtask interface to align VLM and VLA, which should meet the requirements of executability and tractability. Next, we first introduce the formulation and then present how to construct data for training in the subsequent subsections.

\noindent\textbf{Interface Formulation.}
To achieve generalist embodied autonomy, a hierarchical agent must reliably translate open-ended human intentions (the global instruction $I$) into continuous, high-frequency motor commands (the action sequence $a_t$). However, directly mapping unconstrained high-level cognitive reasoning to low-level reactive control inevitably induces a severe semantic-kinematic gap. We resolve this by instantiating the \emph{subtask} ($s_t$) as a bidirectionally aligned interface. For example, given a complex global instruction $I$ such as ``\emph{Wash the beaker}'', the interface grounds the plan into executable skill primitives, such as $s_1$: ``\texttt{[Pick]} \emph{the beaker from the table}''. 

While the interface $s_t$ guarantees immediate kinematic grounding, a system executing isolated subtasks easily trapps in execution loops when visual states appear ambiguous. Therefore, an actively updated semantic memory, $M^{(t)}$, is introduced as the temporal bridge connecting discrete subtasks. At any given timestep $t$ during the $k$-th subtask phase, this temporal prior aggregates all previously achieved milestones:
$M^{(t)} = M^{(0)} \oplus \bigoplus_{i=1}^{k-1} \Phi(s_i)$
where $\oplus$ denotes semantic concatenation and $\Phi(s_i)$ encodes the completed state of the $i$-th subtask. Starting from an initial memory $M^{(0)}$ (e.g., ``\emph{This is the first subtask, and no subtasks have been completed yet}''), $M^{(t)}$ actively aggregates prior milestones into a consolidated history. To illustrate, before generating the 4th subtask ($s_4$: ``\texttt{[Unscrew]} \emph{the cap of the water bottle}''), the memory explicitly logs: ``\emph{The robot has picked up the beaker and placed it on the platform, then picked up the water bottle}''.
\begin{figure}[!t]
\centering
\includegraphics[width=1.0\textwidth]{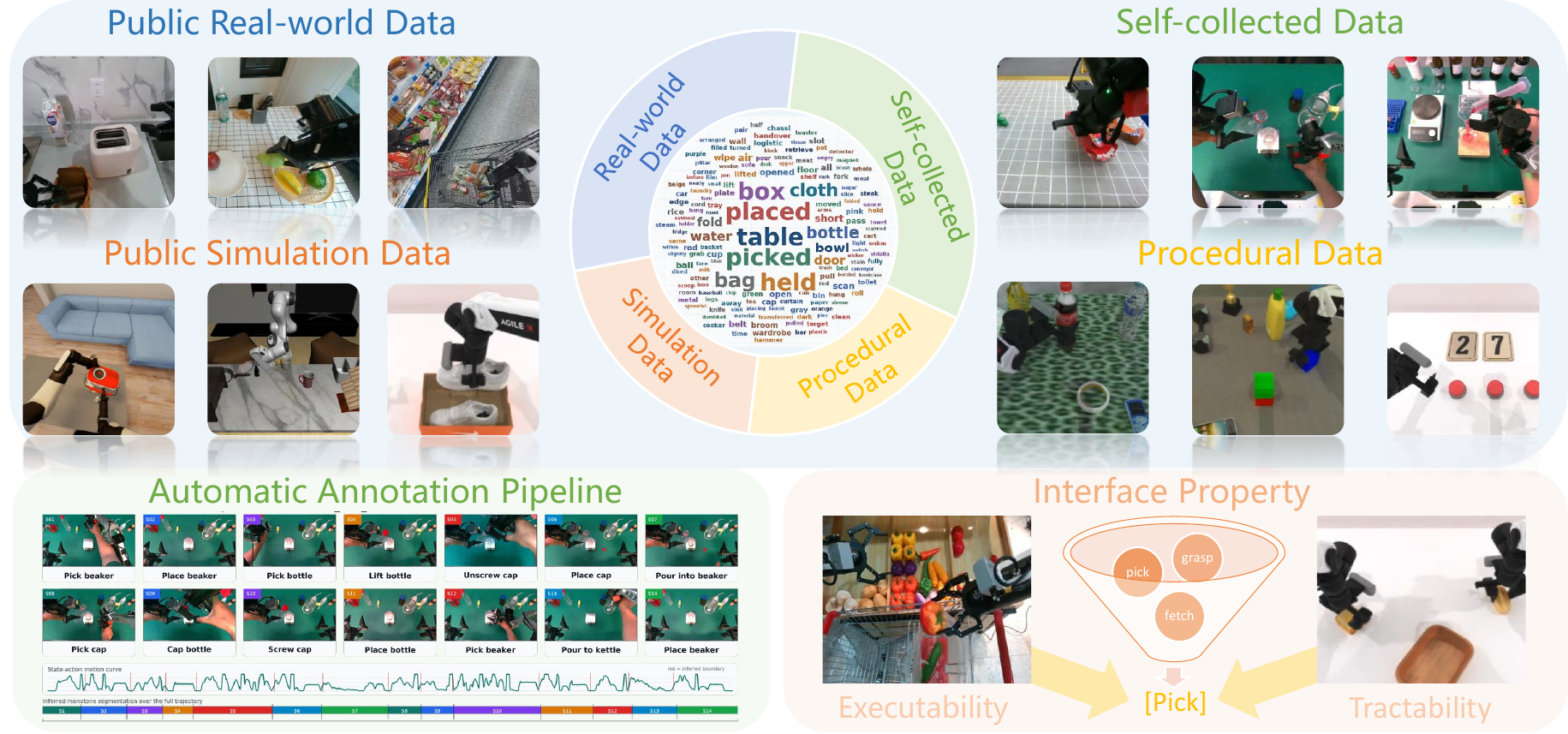}
\caption{Long-horizon metadata construction and interface standardization. Our data generation pipeline leverages automated annotation of over 4,000 hours of open-source video data and synthesizes high-quality procedural simulation data.}
\vspace{-7mm} 
\end{figure}

\subsection{Metadata Annotation \& Generation}
Given the interface definition, this section details the corresponding scalable pipeline used to construct the structural metadata for subsequent training.

\noindent\textbf{Open-source Data Annotation.}
We aggregate over 4,000 hours of long-horizon episodes (with an average of more than $7$ subtasks) from established benchmarks (e.g., AgibotWorld \cite{bu2025agibot,agibotworld2026}, Galaxea \cite{jiang2025galaxea}, BEHAVIOR-1K \cite{li2023behavior}, RoboCerebra \cite{han2026robocerebra}) and newly collected real-world teleoperation data. To ensure \textbf{executability}, we standardize 32 canonical skill primitives. The full definitions of these skill primitives are elaborated in Appendix~\ref{app:skill_vocabulary}. Each primitive is assigned a skill category and a strict language template (e.g., \texttt{[Unscrew]} [object], \texttt{[Stack]} [object1] on [object2]) to establish a uniform syntax. For existing open-source datasets, we utilize Qwen3-VL-235B \cite{Qwen3-VL} to re-annotate subtasks according to our templates and explicitly merge static frames. 
% For real-world data, we deploy a full automated pipeline where Qwen3-VL \cite{Qwen3-VL} generates initial subtask sequences, which are temporally aligned into precise action boundaries using a dynamic programming guided by visual-proprioceptive features. Comprehensive implementation details of this full automated pipeline are provided in Appendix~\ref{app:dataset_details}. For tractability, all annotations are prompted to be enriched with fine-grained descriptions, such as spatial and numerical attributes, enabling the explicit encoding of spatial constraints and interaction counts.

For real-world demonstrations, we deploy an automated boundary inference pipeline that converts raw trajectories into structured subtask annotations. First, coarse boundary priors are generated for reference episodes by prompting Qwen-VL on uniformly sampled frames. For remaining episodes, we frame segmentation as a multimodal sequence partitioning problem. At each frame $t$, we fuse state-action $(o_t, a_t)$ and visual features $I_t$ into a joint representation $x_t = [\phi_s(o_t, a_t), \phi_v(I_t)]$. Dynamic programming is then applied to compute the optimal monotone boundaries $b_1 < \dots < b_{K-1}$ by minimizing the frame-to-subtask compatibility cost, regularized by duration priors and low-motion boundary penalties. 
Comprehensive implementation details are provided in Appendix~\ref{app:dataset_details}.

\noindent\textbf{Simulation Procedural Data Generation.}
% \begin{wrapfigure}{R}{0.3\columnwidth} % 适当加大盒子宽度，比如0.4
%     \centering
%     \includegraphics[width=0.9\linewidth]{image/prompt.pdf} % 占满 wrapfigure 盒子的 95%
%     \caption{Inference detail.}
%     \label{fig:prompt_details}
% \end{wrapfigure}
We utilize simulators, including RoboTwin \cite{chen2025robotwin} and RMBench \cite{chen2026rmbench}, to procedurally synthesize novel, optimal sequences from high-level tasks. We generate subtask annotations online during expert demonstration collection. Subtask boundaries are programmatically defined during expert execution as described in Append \ref{app:robottwin_data}: a scope initiates upon entering a semantic stage and terminates when constituent low-level primitives complete.
To enforce \textbf{tractability}, we extract detailed simulation assets to determine object categories and derive colors directly from their material properties (e.g., "blue toy car"). When multiple identical assets exist within a scene, we access object pose to disambiguate them using relative spatial identifiers (e.g., "the right white stapler"). Interaction counts are annotated using rule-based methods. Crucially, the data generation pipeline explicitly incorporates descriptions of the robot embodiment to evaluate reachability, ensuring the synthesis of optimal subtask routing and kinematically constrained execution paths. 
% Through this pipeline, we re-annotated official benchmark datasets and procedurally generated new episodes, yielding over 30 hours of simulation data. The newly collected data is utilized exclusively for VLA training for fair comparison.

\begin{figure}[!t]
\centering
\includegraphics[width=1.0\textwidth]{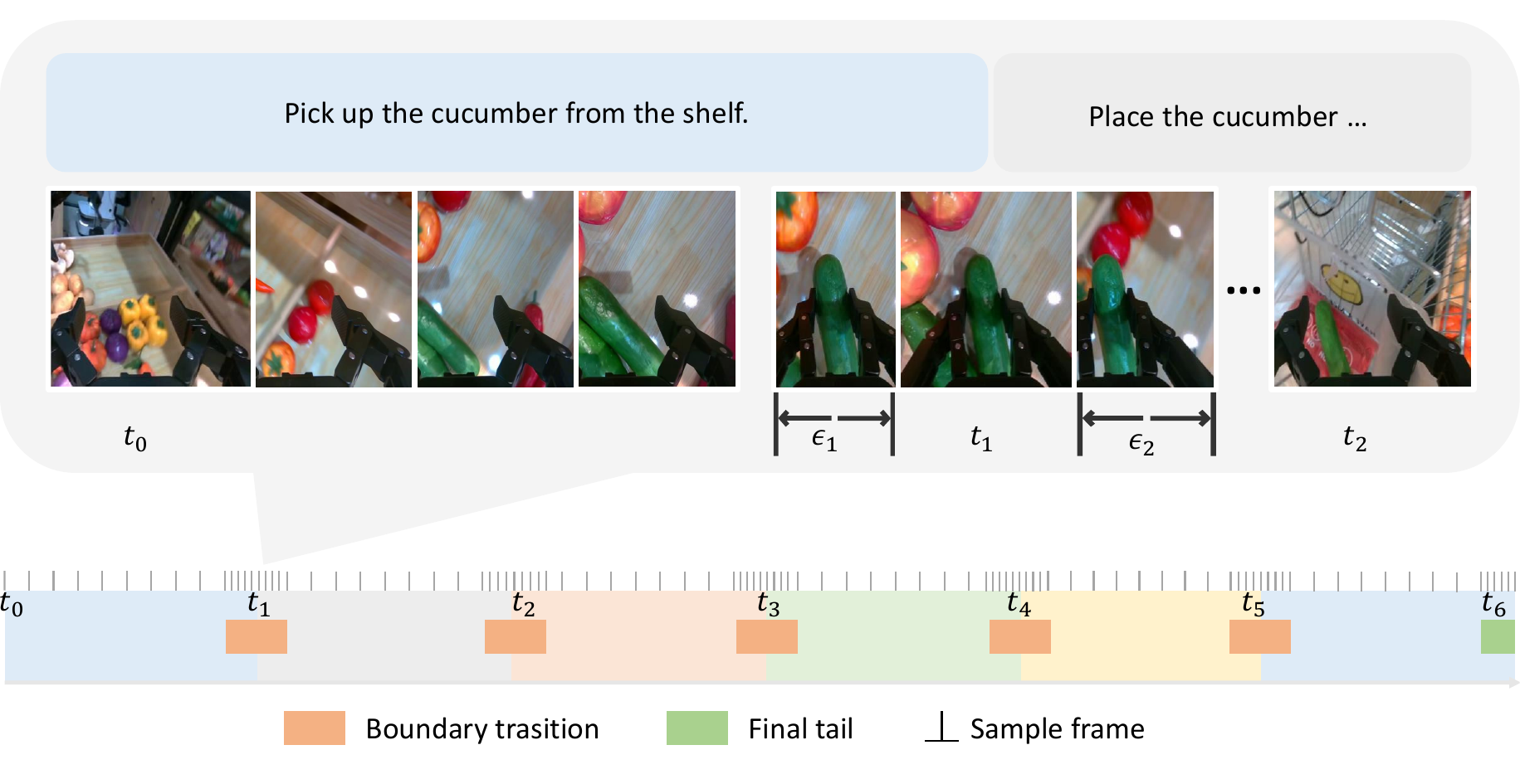}
\vspace{-7mm}
\caption{Event-balanced Sampling. Training trajectories are strategically divided into boundary transition phases and intra-task execution phases. The sampling mechanism employs asymmetric temporal margins around the subtask boundaries, explicitly balancing the learning of continuous state maintenance with discrete logical memory updates.}
\label{fig:architecture}
\vspace{-6mm}
\end{figure}

\subsection{Training}
% Metadata -> how to train -> temporal ambiguity -> event-balanced sampling

\noindent\textbf{Event-balanced Sampling.}
Based on the curated metadata, fine-tuning the dual-system framework requires addressing the planning ambiguity challenge during subtask transitions. Standard uniform frame sampling heavily biases the model toward ongoing execution, obscuring critical task boundaries. Therefore, we devise an event-balanced sampling strategy to construct the training data.

First, we define three distinct temporal phases within a trajectory based on their proximity to ground-truth subtask boundaries $t_k$. To capture state changes, we introduce asymmetric temporal margins, $\epsilon_1$ (pre-boundary) and $\epsilon_2$ (post-boundary). The \textbf{boundary transition phase} is defined as $t \in [t_k-\epsilon_1, t_k+\epsilon_2]$. Because the visual completion landmarks typically emerge slightly after the exact subtask boundary, we empirically set $\epsilon_2 > \epsilon_1$, maintaining the total duration of this transition phase at approximately 1 second. Here, the model must visually verify physical completion to resolve transition ambiguity, thereby triggering discrete state advancement by updating the memory and planning the next subtask $s_{k+1}$. The \textbf{intra-task execution phase} spans the remaining intervals, defined as $t \in (t_{k-1}+\epsilon_2, t_k-\epsilon_1)$. During this phase, observations represent steady actions, and the model must learn ``semantic patience" by maintaining the current active subtask $s_k$ without modifying the memory state. The \textbf{final tail phase}, $t \in [t_K-\epsilon_1, t_K]$, concludes the trajectory. The model processes the final memory state and current visual observations to explicitly output a termination token, signifying that the global instruction has been fully realized.
To balance the learning of continuous state maintenance and discrete logical advancement, we target an near equal sampling ratio between these phases. Given the scarcity of transition and final frames, we employ a denser sampling stride within these boundary phases, dynamically adjusting these strides per dataset to accommodate varying subtask lengths. Specific data ratios are detailed in Appendix~\ref{app:sys2_details}.

We first finetuning on the System-2 VLM \cite{Qwen3-VL} using this event-balanced dataset, optimizing it to actively predict the current subtask and update memory from visual streams. Subsequently, we fine-tune the System-1 VLA \cite{intelligence2025pi05} on the corresponding motor commands, strictly conditioned on the executable subtasks provided by the VLM interface.

\subsection{Inference}
% \begin{wrapfigure}[24]{r}{0.38\textwidth}
%   \centering
%   \vspace{-10pt}
%   \includegraphics[width=\linewidth]{image/prompt.pdf}
%   \vspace{-6pt}
%   \caption{Harness Engineering. The deployment harness acts as a lightweight arbitration layer that maps diverse, open-ended global instructions into a unified instruction interface.}
%   \label{fig:architecture}
%   \vspace{-8pt}
% \end{wrapfigure}

\begin{figure}[!t]
\centering
\includegraphics[width=1.0\textwidth]{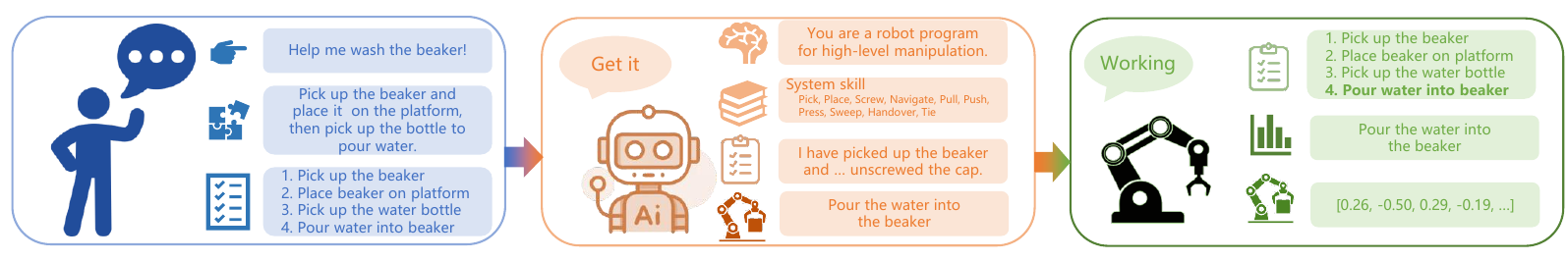}
\caption{Harness Engineering. The deployment harness acts as a lightweight arbitration layer that maps diverse, open-ended global instructions into a unified instruction interface.}
\label{fig:architecture}
\vspace{-5mm} 
\end{figure}

\noindent\textbf{Asynchronous Inference.}
To maximize reactivity without stalling the high-level orchestrator, the dual systems operate continuously and asynchronously. System-2 runs at a lower frequency, processing streaming visual observations alongside the recurrent memory log to monitor progress.

\noindent\textbf{Harness Engineering.}
To operate as a versatile generalist, an agent must dynamically bridge the gap between open-ended human intentions and the kinematically aligned skills. Cortex achieves this by seamlessly mapping a diverse global instruction $I$-spanning \textbf{coarse abstract goals}, \textbf{detailed procedural descriptions}, and \textbf{explicit subtask lists}—into a unified instruction interface. 
The harness constructs rigorous prompts that provide the VLM with rich task contexts, primarily combining the instruction $I$ and the actively updated memory log $M^{(t)}$. To ensure executability, we enforce strict skill constraints on the VLM's output space. During the post-processing stage, the harness utilizes sequence matching to accurately map the generated raw subtask description to the closest matching canonical primitive within the predefined VLA skill library as shown in Appendix \ref{app:real_harness}.

\section{Experiments}
\label{sec:experiments}

Our experimental evaluation is rigorously designed to answer three primary questions, explicitly mapped to the core architectural components of the Cortex framework:
\textbf{(Q1) [Interface \& Metadata]:} How effectively do the curated structural metadata (ensuring executability and tractability) and multi-granular instruction priors ground the cognitive orchestrator against kinematic hallucinations? 
\textbf{(Q2) [Resolving Ambiguities]:} How successfully does Cortex resolve semantic ambiguity (via Spatial/Numerical Grounding) and temporal ambiguity (via Event-balanced Sampling) to surpass monolithic VLAs in long-horizon simulations? 
\textbf{(Q3) [Asynchronous Bidirectional Loop]:} Can the joint asynchronous inference framework reliably transfer to complex real-world environments to achieve dynamic, closed-loop physical error recovery?

\subsection{Open-loop VLM Evaluation}
\label{subsec:exp_top_down}

% To address \textbf{(Q1)} and validate the architectural designs of the cognitive orchestrator, we evaluate the capacity of System 2 to generate planning sequences that preclude kinematic hallucinations and spatial ambiguity. 
% Evaluating generative planning for long-horizon manipulation is notoriously difficult; exact string matching is overly brittle, and traditional multiple-choice benchmarks fail to reflect open-ended generation. Therefore, we employ a flexible \textit{LLM-as-a-Judge} protocol using Qwen-3.5-9B \cite{qwen359b2026}. The judge assesses three core capabilities: \textit{Long-Horizon Logical Consistency}, \textit{Object Counting Accuracy}, and \textit{Spatial Grounding}.
% To bypass the brittleness of literal string matching and the closed-ended limits of multiple-choice formats in evaluating open-ended long-horizon planning, we adopt an \textit{LLM-as-a-Judge} protocol leveraging Qwen-3.5-9B \cite{qwen359b2026} across three core axes aligned with Table \ref{tab:performance_evaluation}:  \textit{Long-Horizon Logical Consistency}, \textit{Object Counting Accuracy}, and \textit{Spatial Grounding}. 

To address \textbf{(Q1)} and validate the interface design, we evaluate System-2's ability to generate physically grounded plans using an \textit{LLM-as-a-Judge} (Qwen-3.5-9B \cite{qwen359b2026}) across three axes: Spatial Information, Long-Horizon Task consistency, and Counting Task accuracy (see Appendix~\ref{app:judge_protocol} for detailed task configurations). We employ two rigorous paradigms: (1) \textbf{Step-Level (Teacher-Forced)} isolates single-step generation accuracy utilizing ground-truth memory, while (2) \textbf{Episode-Level (Self-Forced)} demands autoregressive memory input to measure resilience against compounding semantic drift.

\begin{table*}[t]
\centering
\caption{Performance evaluation on Step-level and Episode-level across different spatial and temporal tasks. Metric scores include Subtask accuracy, Memory (Mem.) accuracy, and Total (Tot.) score.}
\label{tab:performance_evaluation}
\resizebox{\textwidth}{!}{
\begin{tabular}{llccccccccccc}
\toprule
\multirow{2}{*}{\textbf{Level}} & \multirow{2}{*}{\textbf{Method}} & \multirow{2}{*}{\textbf{Avg. Tot.}} & \multicolumn{3}{c}{\textbf{Spatial Information}} & \multicolumn{3}{c}{\textbf{Long-horizon Task}} & \multicolumn{3}{c}{\textbf{Counting Task}} \\
\cmidrule(lr){4-6} \cmidrule(lr){7-9} \cmidrule(lr){10-12}
& & & Subtask & Mem. & Tot. & Subtask & Mem. & Tot. & Subtask & Mem. & Tot. \\
\midrule

% ================= Step-level =================
\multirow{7}{*}{Step-level} 
& Qwen3-VL-8B-Instruct \cite{Qwen3-VL}   & 6.739 & 2.162 & 4.263 & 6.424 & 2.611 & 4.165 & 6.775  & 2.948 & 4.070 & 7.018 \\
& GPT-5 \cite{gpt5systemcard2025}       & 6.268 & 2.755 & 3.668 & 6.422 & 2.481 & 3.682 & 6.163  & 2.790 & 3.429 & 6.220 \\
& Gemini \cite{geminipro31preview2026}  & 6.925 & 2.521 & 4.176 & 6.697 & 2.705 & 4.215 & 6.920  & 3.135 & 4.025 & 7.159 \\
\cmidrule(lr){2-12}
& Cortex (baseline)          & 7.051 & 2.761 & 3.744 & 6.505 & 2.940 & 4.033 & 6.973  & 3.503 & 4.173 & 7.676 \\
& Cortex (w/o harness)          & 7.213 & 2.994 & 3.933 & 6.927 & 3.112 & 3.952 & 7.064  & 3.435 & 4.213 & 7.648 \\
\cmidrule(lr){2-12}
& Cortex (harness on skills)         & 7.392 & 2.885 & 4.150 & 7.035 & 3.144 & 4.121 & 7.265  & 3.527 & 4.348 & 7.875 \\
& \textbf{Cortex (full harness)} & \textbf{8.318} & \textbf{3.800} & \textbf{4.254} & \textbf{8.053} & \textbf{3.909} & \textbf{4.250} & \textbf{8.160}  & \textbf{4.237} & \textbf{4.504} & \textbf{8.741} \\
\midrule

% ================= Episode-level =================
\multirow{7}{*}{Episode-level} 
& Qwen3-VL-8B-Instruct\cite{Qwen3-VL}   & 6.292 & 2.548 & 3.920 & 6.468 & 2.274 & 3.746 & 6.021  & 2.614 & 3.775 & 6.388 \\
& GPT-5\cite{gpt5systemcard2025}       & 7.231 & 2.920 & 4.400 & 7.321 & 2.711 & 4.285 & 6.996  & 3.320 & 4.056 & 7.376 \\
& Gemini \cite{geminipro31preview2026}  & 6.860 & 2.896 & 4.033 & 6.929 & 2.774 & 3.870 & 6.644  & 3.060 & 3.945 & 7.006 \\
\cmidrule(lr){2-12}
& Cortex (baseline)          & 6.978 & 2.549 & 3.969 & 6.518 & 2.238 & 4.362 & 6.600  & 3.525 & 4.290 & 7.815 \\
& Cortex (w/o harness)          & 7.155 & 3.033 & 3.771 & 6.805 & 3.004 & 3.922 & 6.925  & 3.741 & 3.995 & 7.736 \\
\cmidrule(lr){2-12}
& Cortex (harness on skills)           & 7.314 & 3.017 & 3.958 & 6.985 & 2.743 & 4.123 & 6.866  & 3.950 & 4.142 & 8.091 \\
& \textbf{Cortex (full harness)} & \textbf{7.810} & \textbf{3.543} & \textbf{4.044} & \textbf{7.587} & \textbf{3.576} & \textbf{3.804} & \textbf{7.380}  & \textbf{4.203} & \textbf{4.262} & \textbf{8.464} \\

\bottomrule
\end{tabular}
}
\vspace{-1.5\intextsep}
\end{table*}

To evaluate the individual contribution of each component in Cortex, we compare four strategic variants (results detailed in Table~\ref{tab:performance_evaluation}):
% \begin{itemize}
% \item \textbf{Cortex (baseline):} A stripped-down variant relying solely on event-balanced sampling.
% \item \textbf{Cortex (w/o harness):} Complements the baseline with interface information but excludes harness engineering.
% \item \textbf{Cortex (harness on skills):} Integrates interface information and event-balanced sampling, but restricts the harness engineering exclusively to skills.
% \item \textbf{Cortex (full harness):} The complete framework seamlessly incorporating all three proposed modules.
% \end{itemize}
(1) \textbf{Cortex (baseline)}, a stripped-down variant relying solely on event-balanced sampling; 
(2) \textbf{Cortex (w/o harness)}, which complements the baseline with interface information; 
(3) \textbf{Cortex (harness on skills)}, which restricts the harness engineering to skills; and 
(4) \textbf{Cortex (full harness)}, the complete proposed framework.
As shown in Table~\ref{tab:performance_evaluation}, step-by-step introduction of each component yields consistent performance gains, with the \textbf{full harness} configuration achieving the optimal average total scores at both step-level (8.318) and episode-level (7.810).

\begin{wraptable}{r}{0.5\textwidth} 
\vspace{-2.8\intextsep}
% \vspace{-1.5\intextsep}
\centering
\caption{\textbf{Zero-shot} comparison on LIBERO-Long}
\label{tab:libero_long_comparison}
\small
\vspace{-0.1em}
\setlength{\tabcolsep}{4pt}
\begin{tabular}{llc}
\toprule
\textbf{Paradigm} & \textbf{Method} & \textbf{Success (\%)} \\
\midrule
\multirow{4}{*}{\textit{End-to-End}} 
& $\pi_{0}$ \cite{black2024pi0} & 85.2 \\
& $\pi_{0.5}$ \cite{intelligence2025pi05} & 92.4 \\
& MemoryVLA \cite{shi2025memoryvla} & 93.4 \\
& OpenVLA-OFT \cite{kim2025fine} & 94.5 \\
\midrule
\multirow{5}{*}{\textit{Agentic}} 
& RoboBrain \cite{tan2026robobrain} & 57.0 \\
& Qwen3-VL-8B \cite{Qwen3-VL} & 68.0 \\
& GPT-5.4 \cite{gpt54systemcard2026} & 72.0 \\
& Gemini-3.1-Pro \cite{geminipro31preview2026} & 91.0 \\
& \textbf{Cortex (Ours)} & \textbf{95.5 \scriptsize{(+3.1)}} \\
\bottomrule
\end{tabular}
\vspace{-9mm}
\end{wraptable}

\subsection{Closed-loop Simulation Evaluation}
\label{subsec:simulation}

To address \textbf{(Q2)}, we evaluate the complete dual-system in challenging, multi-stage simulation environments. Crucially, since ground-truth subtasks are inaccessible during closed-loop rollouts, all evaluations are strictly conditioned on \textbf{raw, high-level global instructions}.

% This explicitly penalizes open-loop execution and rigorously tests the framework's ability to autonomously resolve semantic and temporal ambiguities end-to-end.

% To address \textbf{(Q2)}, we evaluate how the reactive System-1 executor performs when continuously guided by these standardized subtasks. We deploy Cortex in challenging, multi-stage environments. These benchmarks explicitly test a policy's ability to resist compounding covariate shift and Markovian short-sightedness.

\paragraph{LIBERO-Long Benchmark.} As detailed in Table \ref{tab:libero_long_comparison}, we rigorously isolate reasoning capabilities by coupling all agentic methods (including Cortex) with $\pi_{0.5}$ as their shared System-1 executor. Crucially, Cortex achieves a state-of-the-art \textit{zero-shot} success rate of \textbf{95.5\%}. While end-to-end models like $\pi_{0.5}$ (92.4\%) perform adequately, they operate purely reactively. Leading generalist agents like Gemini-3.1-Pro (91.0\%) act as disembodied observers; lacking strict physical grounding, they occasionally hallucinate subtasks that violate immediate kinematic constraints.

% \paragraph{LIBERO-Long Benchmark.} As detailed in Table \ref{tab:libero_long_comparison}, we categorize baselines into \textit{End-to-End Policies} and \textit{Agentic Frameworks}. To rigorously isolate the reasoning capabilities, all agentic methods (including Cortex) are systematically coupled with $\pi_{0.5}$ as their shared underlying System-1 execution VLA. Crucially, despite System 2 never encountering Libero data during training, Cortex achieves a state-of-the-art \textit{zero-shot} success rate of \textbf{95.5\%}. While monolithic end-to-end models like $\pi_{0.5}$ alone achieve strong results (92.4\%), they operate purely reactively and are prone to short-sightedness. Furthermore, leading generalist agents like Gemini-3.1-Pro (91.0\%) act as disembodied semantic observers lacking bidirectional alignment, occasionally hallucinating subtasks that violate immediate kinematic constraints.

% \begin{wrapfigure}{r}{0.6\columnwidth} 
%     \vspace{-2.5em}
%     \centering
%     \includegraphics[width=0.595\columnwidth]{image/exp_robotwin.pdf} 
%     \vspace{-2.5em}
%     \caption{Success rates on RoboTwin 2.0 benchmark.}
%     % \caption{Success rates on RoboTwin 2.0 under the data-scaling setting.}
%     \label{fig:robotwin_comparison_easy}
% \end{wrapfigure}

\begin{wrapfigure}{r}{0.5\columnwidth} 
    \vspace{-2.5em}
    \centering
    \includegraphics[width=0.5\columnwidth]{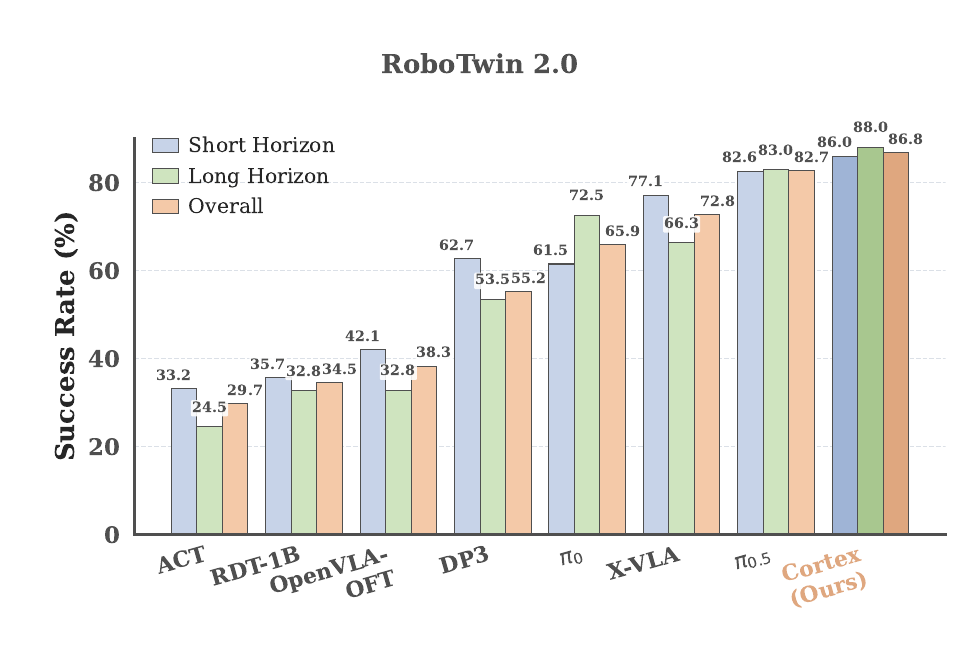} 
    \vspace{-2.5em}
    \caption{Success rates on RoboTwin benchmark.}
    % \caption{Success rates on RoboTwin 2.0 under the data-scaling setting.}
    \label{fig:robotwin_comparison_easy}
    \vspace{-1.0em}
\end{wrapfigure}

\paragraph{RoboTwin Evaluation.} As illustrated in Figure \ref{fig:robotwin_comparison_easy}, monolithic VLAs suffer performance degradation as the task horizon scales, succumbing to premature task completion due to temporal ambiguity. Conversely, Cortex utilizing $\pi_{0.5}$ and maintains an exceptionally high success rate (\textbf{88.00\%}) on long-horizon splits. We attribute this robustness directly to our proposed ambiguity resolution mechanisms: (1) \textbf{Resolving Semantic Ambiguity:} Fine-grained attribute and spatial grounding yield significant improvements in visually dense tasks like \textit{place\_object\_basket} (80\% $\rightarrow$ 85\%), conditioning the underlying VLA to predict kinematically favorable chunking. Cortex actively accounts for reachability constraints, dynamically inserting intermediate ``handover'' subtasks in \textit{dump\_bin\_bigbin} (92\% $\rightarrow$ 98\%). (2) \textbf{Resolving Temporal Ambiguity:} Through event-balanced sampling, System-2 actively dictates the temporal flow rather than passively sleeping, dynamically injecting intermediate subtasks (e.g., \textit{press\_button} task in Table~\ref{RMBench result}) to ensure continuous, uninterrupted physical progression.

\subsection{Closed-loop Real-world Evaluation}
\label{subsec:real_world}

\begin{figure}[htbp]
\centering
\includegraphics[width=1.0\textwidth]{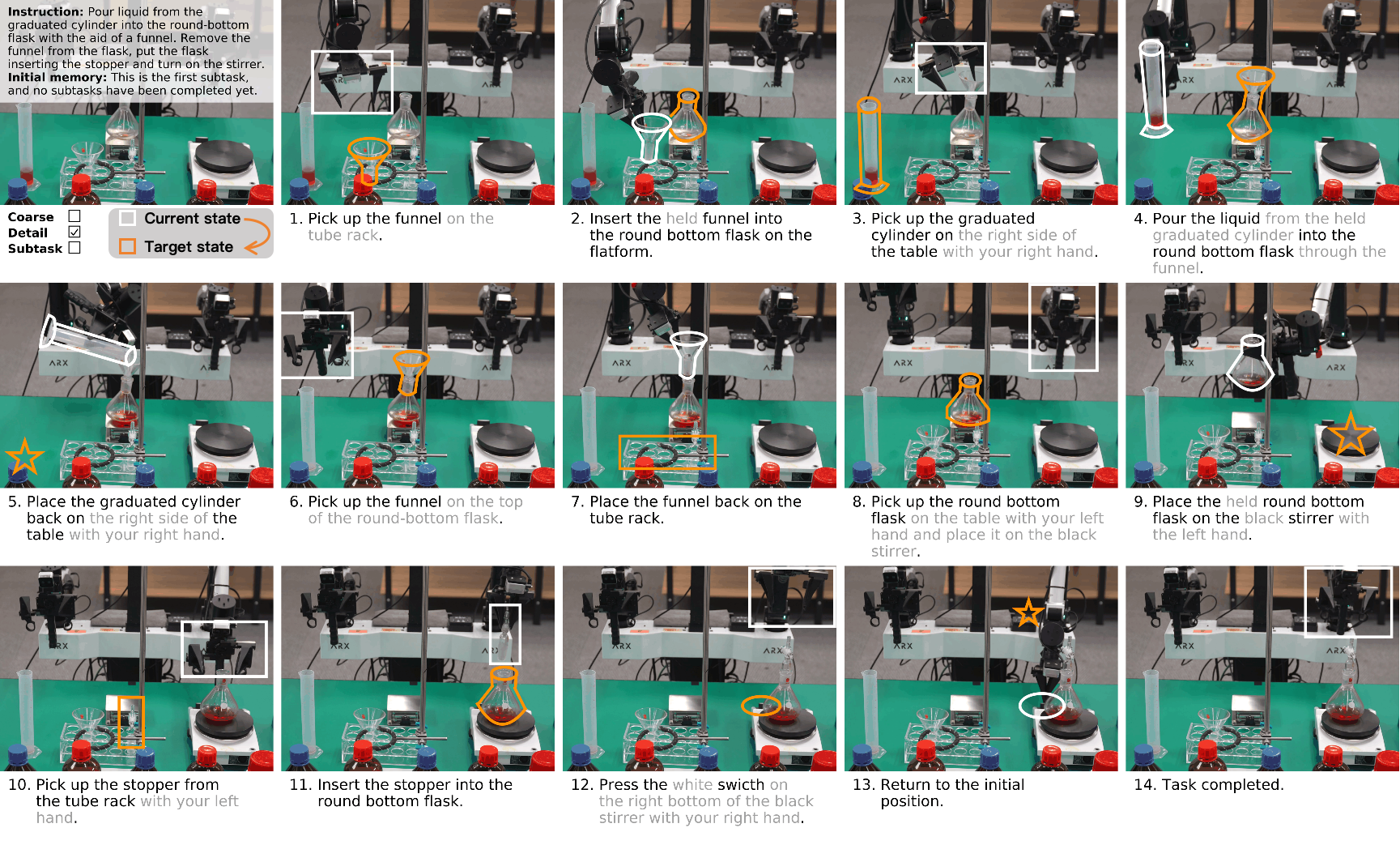}
\caption{Zero-shot Real-world Deployment in Multi-stage Chemistry Tasks. A 14-step continuous experiment demonstrating Cortex's exceptional capability in fine-grained spatial-numerical grounding, robust long-horizon progress tracking, and closed-loop physical execution.}
\label{fig:real_world_14_subtasks}
% \vspace{-5mm} 
\end{figure}

To answer \textbf{(Q3)}, we deploy the Cortex architecture zero-shot on an ARX ACONE physical platform. We design complex, multi-stage chemistry experiments and kitchen assembly workflows that strictly prohibit purely reactive execution. The remarkable zero-shot success of Cortex (Table \ref{tab:memory_vla_results}) stems from the seamless integration of our asynchronous bidirectional loop.

\begin{wraptable}{r}{0.6\linewidth}
\vspace{-3mm}
\centering
\caption{Method comparison across two long-horizon task suites. Progress and success rate are average over 20 trials.}
\label{tab:memory_vla_results}
\vspace{-2mm}
\small
\setlength{\tabcolsep}{3.5pt}
\renewcommand{\arraystretch}{0.92}
\begin{tabular}{llcccc}
\toprule
\multirow{2}{*}{Paradigm} & \multirow{2}{*}{Method}
& \multicolumn{2}{c}{Chemical}
& \multicolumn{2}{c}{Washing} \\
\cmidrule(lr){3-4} \cmidrule(lr){5-6}
& & Prog. $\uparrow$ & SR $\uparrow$
  & Prog. $\uparrow$ & SR $\uparrow$ \\
\midrule
\multirow{2}{*}{End-to-end}
& $\pi_{0.5}$              & 2.5/14  & 0  & 3.7/14  & 0  \\
& $\pi_{\mathrm{mem}}$     & 4.1/14  & 0  & 6.5/14  & 0  \\
\midrule
\multirow{2}{*}{Agentic}
& Cortex                   & 11.0/14 & 65 & 10.5/14 & 55 \\
& Human + $\pi_{\mathrm{mem}}^{\mathrm{sub}}$
                           & \textbf{12.2/14} & \textbf{75}
                           & \textbf{11.6/14} & \textbf{70} \\
\bottomrule
\end{tabular}
\vspace{-3mm}
\end{wraptable}
% yq修改
%To instantiate the reactive executor, System-1 ($\pi_{0.5}$) is fine-tuned on a modest 10-hour dataset curated via our automated subtask segmentation pipeline. While highly capable of executing localized primitives given perfect preceding conditions, a standalone System-1 is inherently brittle to compounding errors across long horizons, inevitably culminating in catastrophic failure as shown in Table \ref{tab:memory_vla_results}. By encapsulating this executor within the Cortex framework, System-2 continuously grounds System-1 via optimally routed, unambiguous subtasks, successfully elevating a short-sighted reactive policy to achieve robust, long-horizon autonomy.

The reactive executor is instantiated as a MEM-style System-1 policy \cite{torne2026mem}, $\pi_{\mathrm{mem}}^{\mathrm{sub}}$, with a several-second memory window. We fine-tune it on about 10 hours of automatically segmented subtask-to-action data. As end-to-end comparisons, we evaluate $\pi_{0.5}$ without explicit memory and a task-level $\pi_{\mathrm{mem}}$ with the same short-memory design. As shown in Table \ref{tab:memory_vla_results}, both remain brittle because they must infer global progress and subtask boundaries from the raw instruction stream, causing repeated primitives and local errors to compound over long horizons. Detailed visual comparisons and failure analyses are provided in Appendix~\ref{app:real_beaker_washing} and Appendix~\ref{app:real_stirring}, with representative cases in Figs.~\ref{fig:app_beaker_e2e_comparison} and \ref{fig:app_chemical_e2e_comparison}. Encapsulating $\pi_{\mathrm{mem}}^{\mathrm{sub}}$ within Cortex lets System-2 continuously ground execution through optimally routed, unambiguous subtasks.

Crucially, the Cortex architecture natively supports \textbf{Closed-Loop Physical Error Recovery}, while a mature \textbf{agent harness} improves both fault tolerance and execution fluency across the asynchronous dual-system interface. Real-world deployment inevitably involves perception-action desynchronization: minor visual occlusions or delayed state transitions can cause reactive policies to stall. Rather than failing passively, Cortex triggers a timeout-driven kinematic reset that introduces a controlled physical perturbation to refresh visual evidence and resolve perceptual deadlocks. The continuously streaming System-2 then immediately re-processes the updated observation, re-validates task progress online, and dispatches the corrected subtask. As a representative real-world case, Figure~\ref{fig:real_world_14_subtasks} shows that Cortex autonomously completes a 14-subtask long-horizon chemistry procedure under this mechanism. More implementation and evaluation details are provided in Appendix~\ref{app:real_robot}.

\section{Conclusion}
This paper introduces Cortex, a bidirectionally aligned dual-system framework designed to overcome the Markovian short-sightedness of monolithic Vision-Language-Action (VLA) models in long-horizon manipulation. To bridge the semantic-kinematic gap, Cortex deeply couples a cognitive orchestrator with a reactive executor via a structured subtask interface, constrained by 32 canonical skill primitives and kinematically optimal routing. This bidirectional alignment synchronizes the dual systems by explicitly resolving both semantic and temporal ambiguities. Consequently, Cortex decisively outperforms purely end-to-end monolithic VLAs in long-horizon simulations, while demonstrating robust zero-shot physical error recovery in complex, real-world chemistry tasks. By formalizing this continuous reasoning-execution loop, Cortex establishes a highly scalable foundation for long-horizon robot autonomy.  
%===============================================================================

\section{Limitations}
Despite its robust planning capabilities, Cortex faces two primary limitations. 
\textbf{Memory Representation:} Text-based memory discards spatial coordinates and visual nuances, disrupting object-instance correspondence during large-scale mobile manipulation. Future work will integrate visual memory retrieval and pixel-level grounding to extend Cortex into a unified dual-mode framework. 
\textbf{High-Frequency States:} Relying on standard vision encoders renders Cortex insensitive to high-frequency micro-state changes. Although we explored tokenizing historical proprioceptive states (analogous to $\pi_{0.5}$) to inject continuous kinematic priors, seamlessly fusing these with vision for rapidly changing, highly dynamic environments remains a formidable challenge.

%===============================================================================

\clearpage
% The acknowledgments are automatically included only in the final and preprint versions of the paper.
% \acknowledgments{If a paper is accepted, the final camera-ready version will (and probably should) include acknowledgments. All acknowledgments go at the end of the paper, including thanks to reviewers who gave useful comments, to colleagues who contributed to the ideas, and to funding agencies and corporate sponsors that provided financial support.}

%===============================================================================

\bibliography{root}  % .bib

\clearpage
\appendix
\section{Appendix}
\label{sec:appendix}

This appendix provides supplementary implementation details, evaluation protocols, and qualitative analyses for Cortex. The additional material is organized to clarify how the proposed bidirectional alignment is instantiated in data construction, model training, simulation evaluation, and real-world deployment.

\subsection{Automatic Annotation Pipeline}
\label{app:dataset_details}

To scale long-horizon supervision without dense human trajectory annotation, we adopt an annotation-free subtask boundary inference pipeline that converts raw robot demonstrations into structured pseudo-subtask annotations. Consider a raw robot trajectory
\begin{equation}
    \tau = \{(o_t, a_t, I_t)\}_{t=1}^{T},
\end{equation}
where $o_t$ denotes the robot state, $a_t$ denotes the action, and $I_t$ denotes the multi-view visual observation at frame $t$. Each task is associated with an ordered task-level subtask schema
\begin{equation}
    \mathcal{S} = (s_1, s_2, \ldots, s_K).
\end{equation}
The objective is to infer a monotone boundary sequence
\begin{equation}
    0 = b_0 < b_1 < \cdots < b_K = T,
\end{equation}
so that each segment $[b_{k-1}, b_k)$ corresponds to the $k$-th subtask $s_k$. In this context, annotation-free denotes the absence of manual frame-level or trajectory-level boundary labels; only the ordered task-level subtask names are assumed to be available, for example from task instructions, program structure, or automatically parsed language descriptions.

\begin{figure*}[h]
\centering
\includegraphics[width=0.98\textwidth]{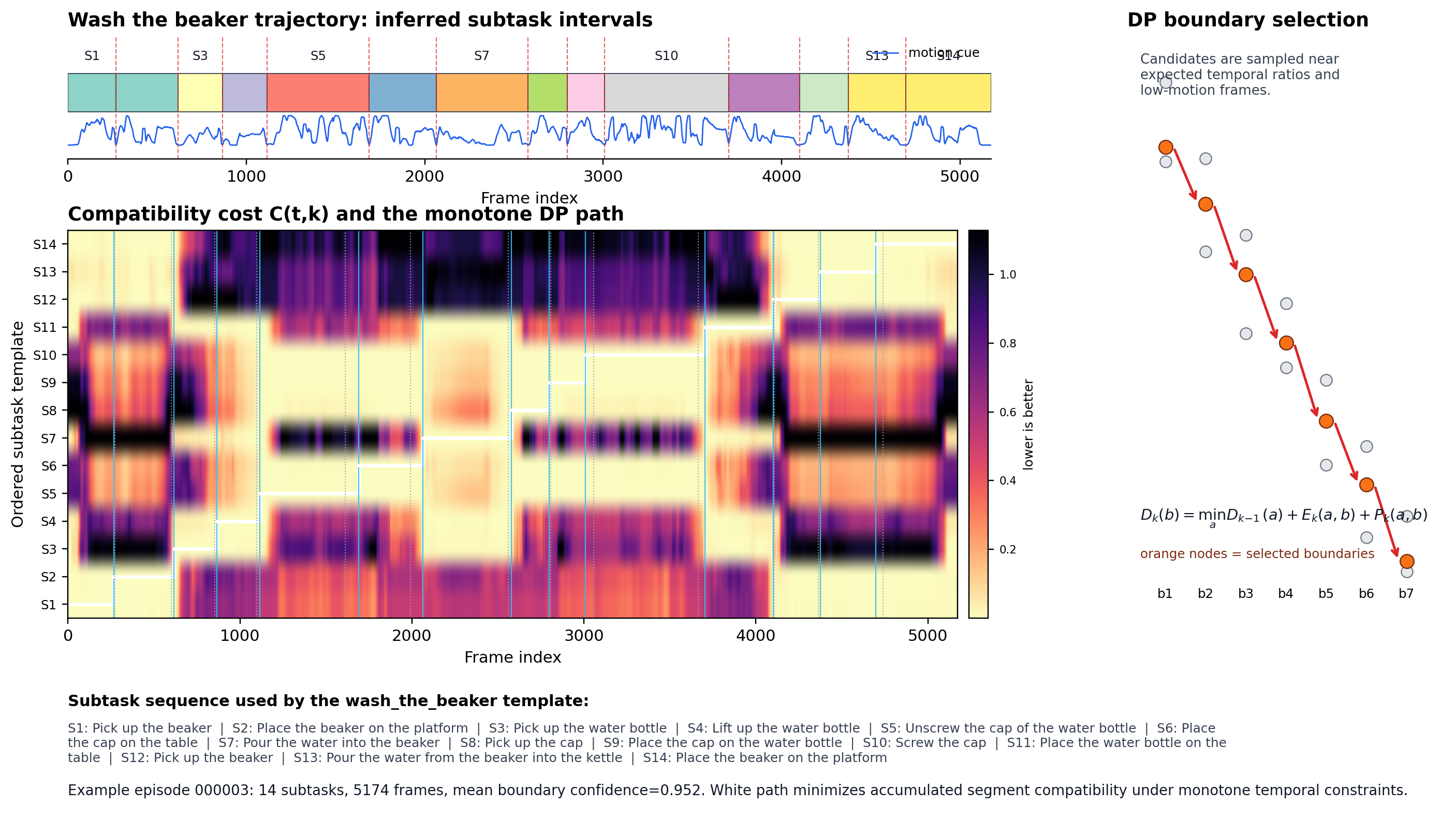}
\caption{Annotation-free subtask boundary inference on \textit{Wash the beaker}. Each thumbnail is sampled at the inferred start boundary of a subtask. The bottom bar shows the monotone temporal segmentation over the full trajectory.}
\label{fig:dataset_annotation_pipeline}
\end{figure*}

\begin{figure*}[t]
\centering
\includegraphics[width=0.98\textwidth]{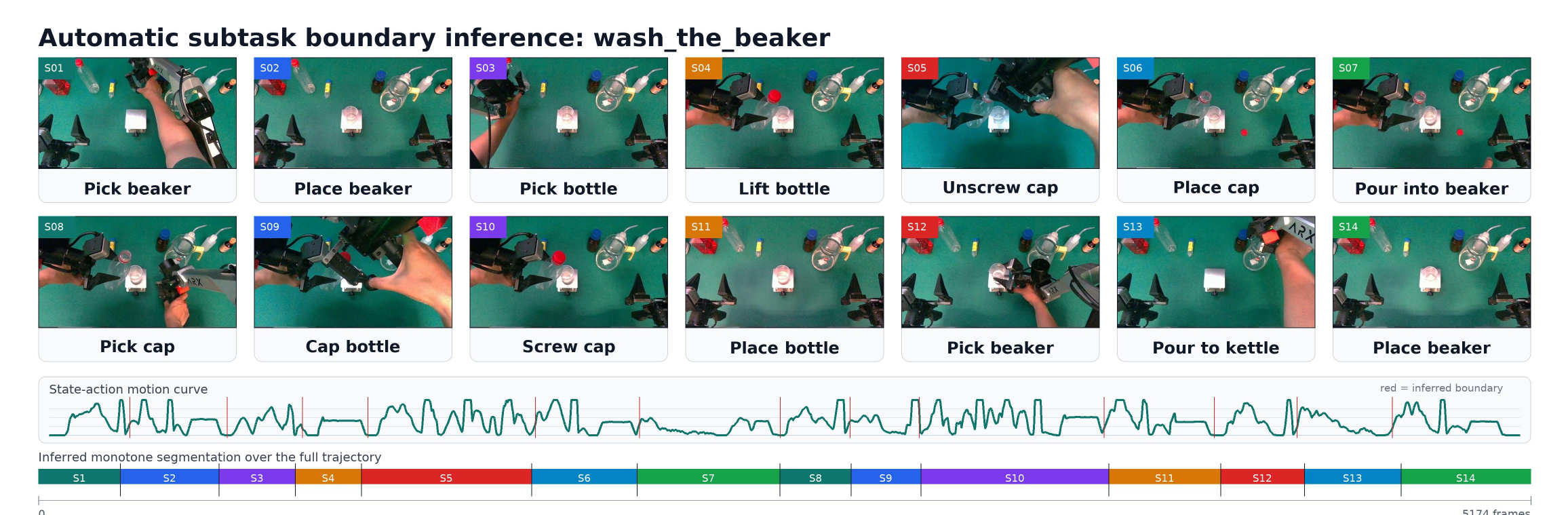}
\caption{Annotation-free boundary inference for a representative \textit{Wash the beaker} trajectory. Thumbnails indicate inferred subtask onsets, and the lower bar shows the resulting monotone segmentation.}
\label{fig:wash_beaker_boundaries}
\end{figure*}

\paragraph{Monotone subtask segmentation.}
Subtask segmentation is formulated as a multimodal sequence partitioning problem. For each frame, we extract state-action and visual features and fuse them as
\begin{equation}
    x_t = \left[
    \sqrt{\lambda_s}\,\phi_s(o_t,a_t),
    \sqrt{\lambda_v}\,\phi_v(I_t)
    \right],
\end{equation}
where $\phi_s$ encodes robot state, action, and local temporal differences, while $\phi_v$ captures downsampled appearance statistics and short-term visual change. The coefficients $\lambda_s$ and $\lambda_v$ balance state and visual modalities. For each subtask $s_k$, we estimate a prototype distribution
\begin{equation}
    p(x_t \mid s_k) = \mathcal{N}(\mu_k, \Sigma_k),
\end{equation}
and define the frame-to-subtask compatibility cost as
\begin{equation}
    c_k(t) = (x_t - \mu_k)^\top \Sigma_k^{-1}(x_t - \mu_k) + \log |\Sigma_k|.
\end{equation}
For a candidate segment $[i,j)$ assigned to subtask $s_k$, the average observation cost is
\begin{equation}
    E_k(i,j) = \frac{1}{j-i} \sum_{t=i}^{j-1} c_k(t).
\end{equation}
To avoid degenerate segmentations, we introduce a duration prior. Let $\rho_k$ denote the expected relative duration of subtask $s_k$; then
\begin{equation}
    P_k(i,j) = \lambda_d \left(\frac{j-i}{T} - \rho_k\right)^2.
\end{equation}
Subtask boundaries also tend to occur near low-motion regions or action-switch points. Let $m_j$ denote the normalized motion magnitude around frame $j$; the boundary penalty is defined as
\begin{equation}
    M(j) = \lambda_m m_j.
\end{equation}
Boundary inference is then performed by dynamic programming over candidate boundary sets $\mathcal{B}_k$, constructed from sampled frames near expected transition locations together with local low-motion minima. The first segment is initialized as
\begin{equation}
    D_1(j) = E_1(0,j) + P_1(0,j) + M(j),
    \qquad j \in \mathcal{B}_1,
\end{equation}
and the recurrence for the $k$-th subtask is
\begin{equation}
    D_k(j) =
    \min_{i \in \mathcal{B}_{k-1},\, i<j}
    \left[
    D_{k-1}(i) + E_k(i,j) + P_k(i,j) + M(j)
    \right],
\end{equation}
for $k=2,\ldots,K-1$, with the terminal segment fixed at $T$:
\begin{equation}
    D_K(T) =
    \min_{i \in \mathcal{B}_{K-1}}
    \left[
    D_{K-1}(i) + E_K(i,T) + P_K(i,T)
    \right].
\end{equation}
Backtracking yields the optimal monotone boundary sequence
\begin{equation}
    b^*_{1:K-1} =
    \arg\min_{b_1 < \cdots < b_{K-1}}
    \sum_{k=1}^{K}
    \left[
    E_k(b_{k-1}, b_k) + P_k(b_{k-1}, b_k)
    \right]
    +
    \sum_{k=1}^{K-1} M(b_k).
\end{equation}
This monotone constraint preserves subtask order and prevents temporal overlap between adjacent segments.

\paragraph{Pseudo-annotation generation.}
Once the optimal boundaries are obtained, each trajectory is converted into a structured subtask annotation containing segment spans, executable action text, and canonical skill labels. A typical record stores, for each subtask, its start frame, end frame, normalized action text, and mapped primitive skill. These pseudo labels are produced entirely by multimodal feature matching and dynamic programming rather than by manual trajectory annotation. During VLA training, the dataloader can therefore replace a sampled frame with its corresponding subtask instruction, such as \textit{pick up the beaker}, \textit{unscrew the cap}, or \textit{pour into the kettle}, thereby providing subtask-conditioned supervision without dense human labeling.

\paragraph{Role in bidirectional alignment.}
This inference pipeline is central to the Cortex Suite because it bridges task-level semantic decomposition and frame-level executable supervision. On the top-down side, it distills heterogeneous free-form demonstrations into standardized subtasks and canonical skills. On the bottom-up side, it provides temporally localized transition labels that are subsequently used to construct ongoing and transition samples for System-2 memory training. Consequently, the same inferred boundary structure supports both kinematic grounding and temporal alignment within a unified data engine.

\subsection{System-2 Training Protocol}
\label{app:sys2_details}

This section details the concrete training configurations, structural data formatting, and empirical validations for the System-2 cognitive planner.

\paragraph{Infrastructure and Hyperparameters.} 
We fine-tune the \texttt{Qwen3-VL-8B-Instruct} architecture utilizing a distributed cluster of 32 NVIDIA A800 (80GB) GPUs. To maximize representation learning capacity, all model parameters—including the vision encoder, multimodal projector, and language model backbone—are fully unfrozen during the optimization process. The entire training trajectory spans approximately 14.2M effective multi-modal samples. For reproducibility, a comprehensive breakdown of the optimization hyperparameters and distributed infrastructure is dynamically tabulated in Table~\ref{tab:system2_pretraining}.

 \begin{table}[h]
\centering
\small
\caption{System-2 training configuration and infrastructure parameters.}
\label{tab:system2_pretraining}
\begin{tabular}{ll}
\toprule
\textbf{Hyperparameter / Item} & \textbf{Configuration} \\
\midrule
Model Backbone & Qwen3-VL-8B-Instruct \\
Hardware Infrastructure & 32 $\times$ NVIDIA A800 GPUs (80GB VRAM) \\
Distributed Engine & DeepSpeed ZeRO-3 \\
Numerical Precision & bfloat16 \\
Optimizer & AdamW \\
Peak Learning Rate & $3\times10^{-6}$ \\
Learning Rate Schedule & Cosine decay \\
Warmup Ratio & 0.03 \\
Weight Decay & 0.0 \\
Global Batch Size & 512 ($32 \times 16$ micro-batch per node) \\
Max Sequence Length & 8192 tokens \\
Visual Resolution Budget & Min pixels: 3,136; Max pixels: 307,200 \\
\bottomrule
\end{tabular}
\end{table}

\paragraph{Unified Instruction Interface.} 
Rather than employing dataset-specific formats, we map heterogeneous supervision sources into a standardized structured prediction problem via a unified prompt template. Concretely, given the \textit{Global Task Goal}, \textit{Input Language Memory}, a fixed 32-way atomic skill vocabulary, and the current observations, the model is trained to generate a structured JSON payload containing \textit{current skill}, \textit{current subtask}, and \textit{active language memory}. For visual inputs, RoboCerebra features a single head-view image, while other datasets supply three synchronized views (head, left wrist, and right wrist). The finalized training mixture reflects a balanced distribution of instruction paradigms, consisting of 42.5\% detailed procedural descriptions, 20.0\% explicit subtask lists, and 37.5\% coarse abstract goals.

\paragraph{Event-Balanced Sampling Implementation and Ablation.}
To ground the event-balanced sampling paradigm, the temporal margin parameter $\epsilon$ is adaptively customized across heterogeneous datasets to accommodate variations in execution pace and annotation granularity. Specifically, $\epsilon$ is decoupled into dataset-specific configurations based on the execution pacing: for rapid, high-velocity tasks (e.g., RoboCerebra), we enforce a tighter margin of $\epsilon = 0.5\text{ s}$; conversely, for slower, more protracted execution sequences (e.g., Galaxea), the window expands up to $\epsilon = 1.5\text{ s}$ to accommodate transitional variance. Through this adaptive stratification, the finalized training corpus exhibits a highly balanced composition between the \textit{intra-task execution phase} and the \textit{boundary transition phase}, shifting from a heavily long-tailed raw distribution to an empirical mixture of approximately 76\% intra-subtask phase samples and 24\% boundary transition phase frames. This deliberate distribution densifies high-fidelity supervisory signals precisely at cognitive pivoting points while ensuring sufficient representation of nominal execution.

To further verify that the observed empirical benefits are inherently attributable to boundary-focused supervision rather than a mere artifact of increased data volume, we conduct a controlled ablation study on the Galaxea \cite{jiang2025galaxea} dataset utilizing a rigorous leave-one-episode-out (\texttt{leave-episode-0-out}) cross-validation protocol. As quantified in Table~\ref{tab:galaxea_uniform_transition_ablation}, our \textit{event-balanced} sampling mixture consistently outpaces the \textit{intra-task-dominant} baseline across all evaluation metrics, despite utilizing fewer total optimization samples (2.72M vs.~3.10M). Specifically, by optimizing the sampling ratio between the intra-task execution phase and the boundary transition phase from 3.77:1 down to a more balanced 2.23:1, the average Subtask accuracy, Memory retention, and Total scores improve from 3.40, 4.17, and 7.58 to 3.85, 4.33, and 8.18, respectively. This substantial gain under a reduced data regime strongly validates our core hypothesis: indiscriminately accumulating redundant steady-state observations within the intra-task execution phase is counterproductive for high-level decision-making. Instead, scaling high-fidelity supervision proximal to the boundary transition phase is paramount, as these critical junctions are precisely where the agent must resolve temporal ambiguity, update its semantic memory, and orchestrate discrete logical advancements.

\begin{table}[t]
\centering
\caption{Ablation study of temporal sample composition on the Galaxea dataset under a \texttt{leave-episode-0-out} validation protocol. Evaluated on the held-out \texttt{episode 0} split across all tasks, the event-balanced mixture confirms superior sample efficiency and task performance. Sample counts are reported in millions (M).}
\label{tab:galaxea_uniform_transition_ablation}
\small
\setlength{\tabcolsep}{4pt}
\renewcommand{\arraystretch}{1.15}
\resizebox{\linewidth}{!}{%
\begin{tabular}{lccccccc}
\toprule
\multirow{2}{*}{\textbf{Sampling Mixture}} & \multicolumn{4}{c}{\textbf{Training Sample Composition}} & \multicolumn{3}{c}{\textbf{Evaluation Score}} \\
\cmidrule(lr){2-5} \cmidrule(lr){6-8}
 & \textbf{Intra-task (M)} & \textbf{Boundary (M)} & \textbf{Total (M)} & \textbf{Ratio (Intra:Bound)} & \textbf{Subtask} & \textbf{Memory} & \textbf{Avg. Total} \\
\midrule
Intra-task-dominant & 2.33 & 0.62 & 3.10 & 3.77:1 & 3.40 & 4.17 & 7.58 \\
Event-balanced (\textbf{Ours}) & 1.78 & 0.80 & 2.72 & 2.23:1 & \textbf{3.85} & \textbf{4.33} & \textbf{8.18} \\
\bottomrule
\end{tabular}%
}
\end{table}

\subsection{LLM-as-a-Judge Protocol \& Benchmark Details}
\label{app:judge_protocol}

To complement Sec.~\ref{subsec:exp_top_down}, we instantiate three evaluation buckets aligned with the three axes in Table \ref{tab:performance_evaluation}: \textit{Spatial Grounding}, \textit{Long-Horizon Logical Consistency}, and \textit{Object Counting Accuracy}. 

\paragraph{Scenario Buckets and Task Allocation.}

\begin{figure*}[t]
\centering
\includegraphics[width=\textwidth]{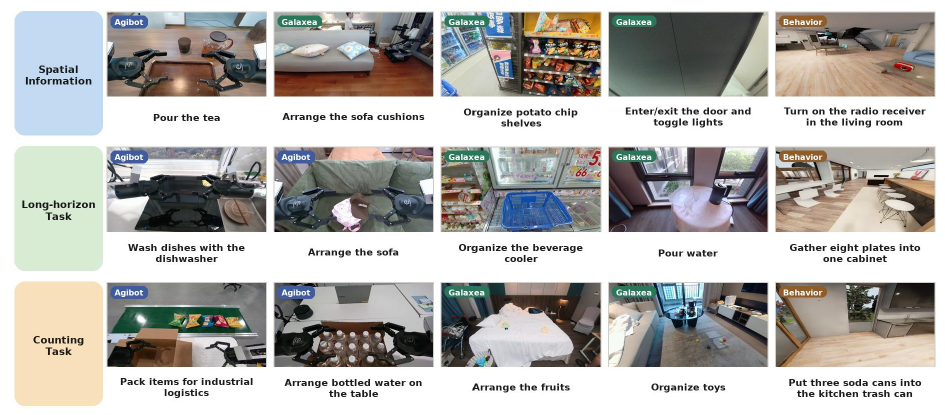}
\caption{Representative raw head-view frames extracted from the 15 tasks in the evaluation suite. The three rows correspond to Spatial Information, Long-horizon Task, and Counting Task, respectively, while the five columns in each row follow the task ordering in Table~\ref{tab:judge_scenario_buckets}.}
\label{fig:judge_bucket_scenarios}
\end{figure*}

To probe complementary facets of System-2 planning, we organize the evaluation suite into three scenario buckets corresponding to spatial grounding, long-horizon sequential reasoning, and counting-sensitive manipulation. Each bucket contains five tasks drawn from AgibotWorld \cite{bu2025agibot}, Galaxea \cite{jiang2025galaxea} and BEHAVIOR-1K \cite{li2023behavior}, yielding 15 rollout tasks in total. Representative scenes for the three buckets are shown in Fig.~\ref{fig:judge_bucket_scenarios}, which may be used to visualize the characteristic observation complexity and planning challenges associated with each evaluation axis.

\begin{table*}[h]
\centering
\caption{Scenario buckets used by the System-2 step-level and episode-level evaluations.}
\label{tab:judge_scenario_buckets}
\small
\setlength{\tabcolsep}{6pt}
\renewcommand{\arraystretch}{1.2}
\begin{tabular}{@{}p{0.19\textwidth}p{0.75\textwidth}@{}}
\toprule
\textbf{Bucket} & \textbf{Evaluation Focus and Source Tasks} \\
\midrule
\parbox[t]{\linewidth}{\raggedright\textbf{Spatial Grounding}} &
\parbox[t]{\linewidth}{\raggedright
\textbf{Primary challenge:} Reference-target disambiguation under distractors, relative position reasoning, and local state verification after manipulation.\par\vspace{2pt}
\textbf{Source tasks:}\\
AgibotWorld: \texttt{task\_598}\\
Galaxea: \texttt{Arrange\_The\_Sofa\_Cushions\_20250722\_008}, \texttt{Organize\_Potato\_Chip\_Shelves\_20250701\_003}, \texttt{Enter\_Exit\_Door\_Turn\_On\_Off\_lights\_20250620\_001}\\
BEHAVIOR-1K: \texttt{task-0000}.} \\
\addlinespace[4pt]
\parbox[t]{\linewidth}{\raggedright\textbf{Long-Horizon Logical Consistency}} &
\parbox[t]{\linewidth}{\raggedright
\textbf{Primary challenge:} Multi-stage causal dependencies, strict execution order, and completion verification before advancing to the next step.\par\vspace{2pt}
\textbf{Source tasks:}\\
AgibotWorld: \texttt{task\_357}, \texttt{task\_527}\\
Galaxea: \texttt{Organize\_The\_Beverage\_Cooler\_20250630\_003}, \texttt{Pour\_Water20250708\_006}\\
BEHAVIOR-1K: \texttt{task-0011}.} \\
\addlinespace[4pt]
\parbox[t]{\linewidth}{\raggedright\textbf{Object Counting Accuracy}} &
\parbox[t]{\linewidth}{\raggedright
\textbf{Primary challenge:} Repeated-object tracking, quantity-sensitive termination conditions, and memory updates driven by count changes.\par\vspace{2pt}
\textbf{Source tasks:}\\
AgibotWorld: \texttt{task\_422}, \texttt{task\_487}\\
Galaxea: \texttt{Arrange\_The\_Fruits\_20250716\_006}, \texttt{Organize\_Toys\_20250628\_002}\\
BEHAVIOR-1K: \texttt{task-0001}.} \\
\bottomrule
\end{tabular}
\end{table*}

For episode-level evaluation, we sample 10 trajectories for each task, yielding 50 trajectories per scenario bucket and 150 trajectories in total. For step-level evaluation, we construct a balanced benchmark from the AgibotWorld \cite{bu2025agibot}, Galaxea \cite{jiang2025galaxea} and BEHAVIOR-1K \cite{li2023behavior} validation splits, yielding approximately 1,000 judged samples for each bucket.

\paragraph{Atomic Skill Vocabulary.}
\label{app:skill_vocabulary}
System-2 does not generate unconstrained free-form actions. Instead, both training and evaluation use a fixed candidate vocabulary of 32 robot atomic skills.

\begin{tcolorbox}[
    breakable,
    colback=gray!8,
    colframe=black!60,
    arc=0mm,
    boxrule=0.4pt,
    top=3mm,
    bottom=3mm,
    left=3mm,
    right=3mm
]
\fontsize{8.5pt}{10.5pt}\selectfont\ttfamily
SYSTEM SKILLS:

Pick, PickAndPlace, Place, Remove, Press, Push, Pull, Navigate, Fold, Wipe, Close, Open, Pour, Cut, Rotate, Handover, Sweep, Stack, Unstack, Screw, Unscrew, Scan, Aim, Clamp, Rinse, Spread, Release, Retreat, AdjustPosture, Tie, Strike, Stir
\end{tcolorbox}

Given the current observation, task instruction, and active language memory, System-2 must first select exactly one skill from this vocabulary when the task is in progress, and then generate the corresponding subtask and updated memory. This constrained skill space reduces open-vocabulary drift and makes the generated subtasks more stable for downstream command normalization and System-1 execution.

\paragraph{Evaluating Prompt Modes.}
\label{app:prompt_modes}
Following the Omni-Format Instruction Interface, we evaluate System 2 under three complementary instruction modalities: \textit{coarse abstract goals}, \textit{detailed procedural descriptions}, and \textit{explicit subtask lists}. These modalities differ in the amount and structure of task-order supervision provided to the planner, while sharing a common prediction objective: conditioned on the current observation and accumulated language memory, the model must identify the currently executable subtask together with the corresponding actively compressed memory state. The unified mode-aware system prompt used across these settings is given below.

\begin{tcbverbatimwrite}{\jobname_sys2_mode_system_prompt.txt}
SYSTEM PROMPT:

You are a robot program for high-level manipulation. Given the global task goal, the input language memory, an optional task-order cue, a list of candidate atomic skills, and the camera observations, first choose exactly one current atomic skill from the candidate skill list when the task is in progress, then predict the subtask the robot should currently be in and the language memory that should be active now.

The optional task-order cue can be either a Detailed Global Task Instruction, a Subtask List, or absent; Detailed Global Task Instruction and Subtask List are mutually exclusive. When a Detailed Global Task Instruction is provided, it is ordered according to the subtask sequence that should be predicted. When a Subtask List is provided, use its item order as the subtask sequence and prefer the corresponding item text for current\_subtask. When no task-order cue is provided, infer the current progress from the global task goal, input language memory, and observations.

Use any provided task-order cue as a constraint when reasoning about progress, and do not skip ahead to later subtasks unless the observations and input language memory clearly indicate that the earlier subtasks have already been completed. If the observations show that the current subtask is still ongoing, keep the same subtask and keep the active language memory unchanged. If the observations show that the previous subtask has just been completed or the robot has already entered the next subtask, predict the next subtask and switch to the new active language memory that reflects the completed progress. If the task has already been completed, set current\_subtask to task\_completed and set current\_skill to null.

Return a JSON object only with keys "current\_skill", "current\_subtask", and "active\_language\_memory". "current\_skill" must be exactly one skill copied from the candidate atomic skill list when the task is in progress, or null when the task is already completed. "active\_language\_memory" should be a concise semantic summary containing only task-relevant completed progress, without low-level visual details or speculation about future subtasks.
\end{tcbverbatimwrite}

\begin{tcolorbox}[
    breakable,
    colback=gray!8,
    colframe=black!60,
    arc=0mm,
    boxrule=0.4pt,
    top=3mm,
    bottom=3mm,
    left=3mm,
    right=3mm
]
\fontsize{8.3pt}{10.4pt}\selectfont\ttfamily
\input{\jobname_sys2_mode_system_prompt.txt}
\end{tcolorbox}

The templates below show the three user-prompt instantiations used consistently in both step-level and episode-level evaluation. They share the same skill-aware prediction format and differ only in how task-order information is supplied to the planner. Fig.~\ref{appx:inference_detail} shows the inference details. 

% \begin{figure}[h]
% \centering
% \includegraphics[width=1.0\textwidth]{image/prompt1.pdf}
% \caption{Inference detail.}
% \vspace{-5mm} 
% \label{appx:inference_detail}
% \end{figure}

\begin{tcbverbatimwrite}{\jobname_sys2_mode_goal_prompt.txt}
Mode 1: Coarse abstract goals

User\_Input = \{ \\
  Global Task Goal: {task\_text}, \\
  Input Language Memory: {input\_language\_memory}, \\
  Candidate Atomic Skills: [skill\_1, ..., skill\_32], \\
  Current Observation Images (in order): \\
    Image 1 ({view\_1}): [image\_1] \\
    ... \\
    Image N ({view\_N}): [image\_N] \\
  Choose exactly one atomic skill from the candidate list when the task is in progress; if the task is already completed, set current\_skill to null and current\_subtask to task\_completed. Predict the subtask and active language memory that should be active now. Return JSON only with keys "current\_skill", "current\_subtask", and "active\_language\_memory". \\
\}
\end{tcbverbatimwrite}

\begin{tcolorbox}[
    breakable,
    colback=gray!8,
    colframe=black!60,
    arc=0mm,
    boxrule=0.4pt,
    top=3mm,
    bottom=3mm,
    left=3mm,
    right=3mm
]
\fontsize{8.3pt}{10.4pt}\selectfont\ttfamily
\input{\jobname_sys2_mode_goal_prompt.txt}
\end{tcolorbox}

\begin{tcbverbatimwrite}{\jobname_sys2_mode_detailed_prompt.txt}
Mode 2: Detailed procedural description

User\_Input = \{ \\
  Global Task Goal: {task\_text}, \\
  Input Language Memory: {input\_language\_memory}, \\
  Detailed Global Task Instruction: {detailed\_task\_text}, \\
  Candidate Atomic Skills: [skill\_1, ..., skill\_32], \\
  Current Observation Images (in order): \\
    Image 1 ({view\_1}): [image\_1] \\
    ... \\
    Image N ({view\_N}): [image\_N] \\
  Choose exactly one atomic skill from the candidate list when the task is in progress; if the task is already completed, set current\_skill to null and current\_subtask to task\_completed. Predict the subtask and active language memory that should be active now. Return JSON only with keys "current\_skill", "current\_subtask", and "active\_language\_memory". \\
\}
\end{tcbverbatimwrite}

\begin{tcolorbox}[
    breakable,
    colback=gray!8,
    colframe=black!60,
    arc=0mm,
    boxrule=0.4pt,
    top=3mm,
    bottom=3mm,
    left=3mm,
    right=3mm
]
\fontsize{8.3pt}{10.4pt}\selectfont\ttfamily
\input{\jobname_sys2_mode_detailed_prompt.txt}
\end{tcolorbox}

\begin{tcbverbatimwrite}{\jobname_sys2_mode_subtask_prompt.txt}
Mode 3: Explicit subtask list

User\_Input = \{ \\
  Global Task Goal: {task\_text}, \\
  Input Language Memory: {input\_language\_memory}, \\
  Subtask List: \\
    1. {subtask\_1} \\
    2. {subtask\_2} \\
    ... \\
    K. {subtask\_K} \\
  Candidate Atomic Skills: [skill\_1, ..., skill\_32], \\
  Current Observation Images (in order): \\
    Image 1 ({view\_1}): [image\_1] \\
    ... \\
    Image N ({view\_N}): [image\_N] \\
  Choose exactly one atomic skill from the candidate list when the task is in progress; if the task is already completed, set current\_skill to null and current\_subtask to task\_completed. Predict the subtask and active language memory that should be active now. Return JSON only with keys "current\_skill", "current\_subtask", and "active\_language\_memory". \\
\}
\end{tcbverbatimwrite}

\begin{tcolorbox}[
    breakable,
    colback=gray!8,
    colframe=black!60,
    arc=0mm,
    boxrule=0.4pt,
    top=3mm,
    bottom=3mm,
    left=3mm,
    right=3mm
]
\fontsize{8.3pt}{10.4pt}\selectfont\ttfamily
\input{\jobname_sys2_mode_subtask_prompt.txt}
\end{tcolorbox}

\paragraph{Qwen-3.5-9B Judge Prompt.}

We use Qwen-3.5-9B \cite{qwen359b2026} as the LLM judge for scoring. For each sample, the judge assigns discrete normalized ratings in $\{0, 0.4, 0.9, 1.0\}$ to subtask correctness and memory correctness, which are subsequently mapped to the reporting scale used in Table \ref{tab:performance_evaluation}, yielding per-axis scores out of 5 and aggregate scores out of 10. To prevent exact lexical matches from being under-scored by the judge, predictions whose subtask or memory strings are character-identical to the reference are deterministically assigned full credit. The judge prompts are listed below:

\begin{tcbverbatimwrite}{\jobname_qwen_judge_system_prompt.txt}
JUDGE\_SYSTEM\_PROMPT:

You are a strict JSON-only judge. Never output thinking process, analysis, or markdown. Return exactly one JSON object and nothing else.
\end{tcbverbatimwrite}

\begin{tcolorbox}[
    breakable,
    colback=gray!8,
    colframe=black!60,
    arc=0mm,
    boxrule=0.4pt,
    top=3mm,
    bottom=3mm,
    left=3mm,
    right=3mm
]
\fontsize{8.3pt}{10.4pt}\selectfont\ttfamily
\input{\jobname_qwen_judge_system_prompt.txt}
\end{tcolorbox}

\begin{tcbverbatimwrite}{\jobname_qwen_judge_user_prompt.txt}
JUDGE USER PROMPT:

You are a strict evaluator.
Do NOT output any thinking process, analysis, explanation, markdown, or extra text.
Output exactly one JSON object only.
Allowed JSON keys: subtask\_score, memory\_score, total\_score, verdict, reason.

Scoring rules:

- First normalize wording before scoring:

  * Pronouns are equivalent for the acting agent (e.g., I/me/my/you/your) and can be ignored when meaning is unchanged.
  
  * If the subject is the robot, pronouns and explicit robot mentions are equivalent (this can still be exact match), and optional robot descriptors are non-essential.
  
  * Treat hand, arm, and gripper as equivalent references to the manipulator.
  
- Use only these discrete scores: 0, 0.4, 0.9, 1.

- subtask\_score rubric:

  * 1: exact match with reference after applying the normalization above. Exact match MUST be 1.0 (never 0.9).
  
  * 0.9: semantically equivalent overall but not exact after normalization (e.g., synonyms, different word order, minor non-critical attribute omission, or non-contradictory object-attribute addition/omission).
  
  * 0.4: describes essentially the same high-level thing but is incomplete or substantially rephrased without explicit contradiction. For multi-part subtasks, predicting only the earlier/front part while missing later parts should be 0.4.
  
  * 0: any logical or factual error exists (wrong action/object/state/relation
  /attribute, missing required action/object, contradiction, or unsupported extra claim). For multi-part subtasks, predicting only a later part while missing the earlier/front part is a logical error $\geq$ 0.
  
- memory\_score rubric:

  * 1 and 0.9: same criteria as subtask\_score.
  
  * 0.4: memory is broadly related but substantially rephrased/coarse, with no contradiction and no omitted required subtask.
  
  * 0: any logical/factual error, contradiction, unsupported extra claim, or omission of any required subtask. Omission $\geq$ 0  (not 0.4).
  
- total\_score = subtask\_score + memory\_score.

- verdict: correct if total\_score=2, partial if 0 $\textless$ total\_score $\textless$ 2, wrong if total\_score=0.

Keep reason concise ($\leq$ 40 words).

[Question] \\
Task Goal: {task\_text} \\
Input Language Memory: {input\_language\_memory} \\

[Reference Answer] \\
{"current\_subtask": "{gt\_current\_subtask}",\\
"active\_language\_memory": "{gt\_active\_language\_memory}"}\\

[Assistant Answer] \\
{pred\_text}

Now output JSON only.
\end{tcbverbatimwrite}

\begin{tcolorbox}[
    breakable,
    colback=gray!8,
    colframe=black!60,
    arc=0mm,
    boxrule=0.4pt,
    top=3mm,
    bottom=3mm,
    left=3mm,
    right=3mm
]
\fontsize{8.3pt}{10.4pt}\selectfont\ttfamily
\input{\jobname_qwen_judge_user_prompt.txt}
\end{tcolorbox}

\subsection{Simulation Evaluation Protocol}
\label{app:simulation_protocol}
\paragraph{Implementation Details.}
  For all simulation benchmarks, we conduct full-parameter finetuning for $\pi_{0.5}$ model as the System-1 executor. The detailed fine-tuning hyperparameters are summarized in the following table \ref{tab:simulation_training}.

  \begin{table}[h]
    \centering
    \small
    \caption{Simulation $\pi_{0.5}$ executor training configuration.}
    \label{tab:simulation_training}
    \begin{tabular}{ll}
    \toprule
    \textbf{Item} & \textbf{Configuration} \\
    \midrule
    Model & Pi05 \\
    Model size & 3.62B parameters \\
    Backbone & PaliGemma, width 2048, depth 18 \\
    Action expert & width 1024, depth 18 \\
    Distributed training & DDP \\
    Precision & bfloat16 \\
    Optimizer & AdamW, $\beta=(0.9,0.95)$, $\epsilon=10^{-8}$ \\
    Learning rate & $5\times10^{-5}$ \\
    LR schedule & Cosine decay with min LR $5\times10^{-6}$ \\
    Warmup steps & 2,000 \\
    Weight decay & $1\times10^{-10}$ \\
    Global batch size & $8 \times 8 = 64$ \\
    Training step & 30, 000 \\
    Max text length & 200 tokens \\
    \bottomrule
    \end{tabular}
  \end{table}
  
\paragraph{LIBERO-Long.}
For LIBERO-Long, all agentic baselines are coupled with the same System-1 executor when feasible, isolating the contribution of high-level planning. A rollout is counted as successful only if the full long-horizon instruction is completed without violating the benchmark success predicate. Intermediate subtask correctness is not directly rewarded unless it contributes to final task success.

\paragraph{RoboTwin.}
\label{app:robottwin_data}
\newcommand{\code}[1]{\texttt{#1}}
For procdural data generation, subtasks come from the expert script itself. Each task file defines an expert procedure in its \code{play\_once()} function. We manually wrap consecutive expert commands with \code{record\_subtask}. Each wrapped block corresponds to one semantic part of the task.
For example, in a bottle-to-dustbin task, the expert script contains blocks of the following form:

\begin{tcolorbox}[
    breakable,
    colback=gray!8,
    colframe=black!60,
    arc=0mm,
    boxrule=0.4pt,
    top=3mm,
    bottom=3mm,
    left=3mm,
    right=3mm
]
\begin{lstlisting}[
    basicstyle=\ttfamily,
    breaklines=true,
    breakindent=1em,
    columns=fullflexible
]
with self.record_subtask("pick up the cola"):
    self.move(self.grasp_actor(bottle, arm_tag=arm_tag, pre_grasp_dis=0.1))
    self.move(self.move_by_displacement(arm_tag, z=0.1))

with self.record_subtask("place the cola into dustbin"):
    self.move((ArmTag("left"), [left_end_action]))
    self.move(self.open_gripper("left"))
\end{lstlisting}
\end{tcolorbox}

The text passed to \code{record\_subtask} becomes the raw subtask description. The commands inside the block define the time span of that subtask. During data generation, the switching signal is the control flow of the expert script. More specifically, a subtask starts when the program enters a \code{record\_subtask} block, and it ends when that block exits.
The implementation is a Python context manager. When the block is entered, the current simulator frame index is saved as \code{start\_frame}. When all commands inside the block finish and the program leaves the block, the current frame index is saved as \code{end\_frame}. The next subtask begins when the program enters the next \code{record\_subtask} block.

RoboTwin evaluation follows the data-scaling protocol described in the main paper. We report success rates on short-horizon tasks, long-horizon tasks, and the overall benchmark. In RoboTwin, we evaluate the proposed dual-system architecture using an online hierarchical control protocol that decouples semantic task progress estimation from low-level action generation. For each evaluation episode, System-1 operates as a vision-language-action policy server that produces short-horizon action chunks conditioned on the currently committed subtask instruction, while System-2 runs as an independent semantic scheduler that periodically infers the active subtask and updates a language memory from multi-view observations and   recent robot state history. In the no-subtask-list setting, System-2 is not provided with the ground-truth ordered subtask sequence; instead, it receives only the global task goal, the previously committed language memory, historical visual observations, and proprioceptive state traces, and outputs a free-form prediction of the current subtask together with an updated memory. To prevent unstable frame-wise semantic predictions from directly perturbing the controller, all System-2 outputs are first processed by a local scheduler, which matches the predicted subtask against an episode-specific subtask plan maintained only on the evaluator side and accepts a transition only when the match confidence, maximum allowed subtask advance, and minimum dwell-time constraints are satisfied. System-1 therefore never consumes raw System-2 predictions directly; it is re-invoked only after the local scheduler commits a new subtask context. This asynchronous evaluation design allows System-1 to execute continuous action chunks at a high control frequency, while System-2 performs lower-frequency semantic monitoring and subtask switching, providing a stable interface between long-horizon task reasoning and closed-loop robotic control.

\begin{table}[htbp]
\centering
\caption{Success rates (\%) on the RoboTwin 2.0 benchmark under the data-scaling setting (50 clean + 500 randomized demonstrations).}
\label{tab:robotwin_comparison_easy}

\footnotesize 
\setlength{\tabcolsep}{6pt} 

\begin{tabular}{lccc}
\toprule
\textbf{Method} & \textbf{short horizon} & \textbf{long horizon} & \textbf{RoboTwin (Overall)} \\
\midrule
ACT \cite{zhao2023learning}                        & 33.20 & 24.50 & 29.70 \\
RDT-1B \cite{liu2025rdt}                    & 35.73 & 32.75 & 34.50 \\
OpenVLA-OFT \cite{kim2025fine}              & 42.10 & 32.80 & 38.30 \\
DP3 \cite{ze20243d}                       & 62.67 & 53.45 & 55.24 \\
$\pi_0$ \cite{black2024pi0}                  & 61.50 & 72.55 & 65.92 \\
X-VLA \cite{zheng2025x}& 77.13 & 66.30 & 72.80 \\
$\pi_{0.5}$ \cite{intelligence2025pi05}              & 82.60 & 82.95 & 82.74 \\
\midrule
\textbf{Cortex (Ours)}           & \textbf{86.00} & \textbf{88.00} & \textbf{86.80(+)} \\
\bottomrule
\end{tabular}
\end{table}

\paragraph{RMBench.}
RMBench emphasizes task memory complexity. We therefore evaluate not only physical success but also whether the generated subtask sequence preserves task-relevant memory variables such as object order, object count, and previously modified states. This setting is particularly useful for diagnosing failures caused by semantic drift rather than low-level motor errors.

\begin{table}[ht]
\centering
\footnotesize 
\renewcommand{\arraystretch}{1.3} % 稍微增加行高，让换行后的表头不显得拥挤
\setlength{\tabcolsep}{3pt} 
\caption{RMBench benchmark results. We report success rates across seven manipulation tasks for six policies, each trained with 50 synthesized demonstrations and evaluated over 100 rollouts.}
\begin{tabularx}{\linewidth}{l *{7}{>{\centering\arraybackslash}X}}
\toprule
\textbf{Method} & \textbf{Observe and Pick Up} & \textbf{Rearrange Blocks} & \textbf{Put Back Block} & \textbf{Swap Blocks} & \textbf{Swap T} & \textbf{Battery Try} & \textbf{Press Button} \\
\midrule
DP \cite{chi2025diffusion}          & 1\%  & 0\%   & 0\%   & 11\% & 20\% & 10\% & 0\%  \\
ACT \cite{zhao2023learning}         & 1\%  & 29\%  & 0\%   & 2\%  & 2\%  & 19\% & 0\%  \\
$\pi_{0.5}$ \cite{intelligence2025pi05}   & 9\%  & 13\%  & 11\%  & 24\% & 15\% & 16\% & 0\%  \\
X-VLA \cite{zheng2025x}             & 9\%  & 13\%  & 18\%  & 16\% & 3\%  & 26\% & 0\%  \\
Mem-0 \cite{chen2026rmbench}        & 4\%  & 89\%  & 90\%  & 67\% & 14\% & 28\% & 0\%  \\
\midrule
\textbf{Cortex (Ours)}              & \textbf{14\%}& \textbf{100\%}& \textbf{100\%}& \textbf{99\%} & \textbf{63\%}& \textbf{37\%}& \textbf{20\%} \\
\bottomrule
\end{tabularx}
% \vspace{-4mm} 
\label{RMBench result}
\end{table}

\paragraph{State history as text.}
Some tasks require high-frequency progress estimation that is not reliably
recoverable from sparse image observations alone. For example, the robot may
need to infer how many times a button has been pressed or whether a handover has
already occurred. We therefore inject a causal history of robot states as text.
For each sample, we read the most recent $H=30$ state slots with stride 1 from
\texttt{observation.state}. We truncate the raw state to at most
$D_{\max}=32$ dimensions; lower-dimensional states keep their native
dimensionality. Each retained scalar is then converted to a bounded integer
token:
\begin{align}
  \bar{s}_{i,d} &= \tanh(s_{i,d}), \\
  b_{i,d} &=
  \mathrm{clip}\left(
    \left\lfloor
    \frac{B}{2}(\bar{s}_{i,d}+1)
    \right\rfloor,\;
    0,\; B-1
  \right),
  \qquad B=256.
  \label{eq:state_quantization}
\end{align}
Missing history rows are represented by the token value $-1$. The prompt
contains the quantized sequence from oldest to current, together with summary
features including per-step $L_1$ changes, the number of changed dimensions, the
oldest-to-current delta, and whether the most recent transition changed. This
textual state representation is intentionally model-agnostic: it can be parsed
by the language backbone without adding a new numeric encoder.

\begin{figure}[t]
  \centering
  \begin{tikzpicture}
    \begin{axis}[
      width=0.96\linewidth,
      height=4.2cm,
      xlabel={time in episode 0 of \texttt{press\_button} (s)},
      ylabel={normalized value},
      xmin=0, xmax=19.2,
      ymin=0, ymax=1.08,
      legend style={font=\scriptsize, at={(0.02,0.98)}, anchor=north west},
      tick label style={font=\scriptsize},
      label style={font=\scriptsize},
      grid=major,
      grid style={line width=.1pt, draw=gray!20},
    ]
      \addplot+[blue, thick, mark=none] coordinates {
        (0.00,0.787) (0.33,0.942) (0.67,0.771) (1.00,0.231)
        (1.33,0.231) (1.67,0.203) (2.00,0.194) (2.33,0.205)
        (2.67,0.000) (3.00,0.000) (3.33,0.203) (3.67,0.204)
        (4.00,0.389) (4.33,0.592) (4.67,0.000) (5.00,0.000)
        (5.33,0.174) (5.67,0.188) (6.00,0.186) (6.33,0.000)
        (6.67,0.000) (7.00,0.185) (7.33,0.187) (7.67,0.031)
        (8.00,0.000) (8.33,0.002) (8.67,0.184) (9.00,0.185)
        (9.33,0.031) (9.67,0.000) (10.00,0.092) (10.33,0.183)
        (10.67,0.185) (11.00,0.005) (11.33,0.000) (11.67,0.183)
        (12.00,0.186) (12.33,0.177) (12.67,0.000) (13.00,0.000)
        (13.33,0.175) (13.67,0.183) (14.00,0.146) (14.33,0.000)
        (14.67,0.000) (15.00,0.177) (15.33,0.182) (15.67,0.458)
        (16.00,0.998) (16.33,1.000) (16.67,0.526) (17.00,0.833)
        (17.33,0.937) (17.67,0.715) (18.00,0.231) (18.33,0.231)
        (18.67,0.211) (19.00,0.145)
      };
      \addlegendentry{state motion, $\sum_d |s_{t,d}-s_{t-1,d}|$}
      \addplot+[red, const plot, thick, mark=none] coordinates {
        (0.00,0.0) (2.53,0.1) (4.17,0.2) (6.17,0.3)
        (7.73,0.4) (9.33,0.5) (10.93,0.6) (12.53,0.7)
        (14.17,0.8) (15.80,0.9) (19.13,1.0)
      };
      \addlegendentry{completed press count}
      \addplot+[gray, dashed, mark=none, forget plot] coordinates {(2.53,0) (2.53,1.05)};
      \addplot+[gray, dashed, mark=none, forget plot] coordinates {(4.17,0) (4.17,1.05)};
      \addplot+[gray, dashed, mark=none, forget plot] coordinates {(6.17,0) (6.17,1.05)};
      \addplot+[gray, dashed, mark=none, forget plot] coordinates {(7.73,0) (7.73,1.05)};
      \addplot+[gray, dashed, mark=none, forget plot] coordinates {(9.33,0) (9.33,1.05)};
      \addplot+[gray, dashed, mark=none, forget plot] coordinates {(10.93,0) (10.93,1.05)};
      \addplot+[gray, dashed, mark=none, forget plot] coordinates {(12.53,0) (12.53,1.05)};
      \addplot+[gray, dashed, mark=none, forget plot] coordinates {(14.17,0) (14.17,1.05)};
      \addplot+[gray, dashed, mark=none, forget plot] coordinates {(15.80,0) (15.80,1.05)};
    \end{axis}
  \end{tikzpicture}
  \caption{State history exposes high-frequency progress in a representative
  \texttt{press\_button} episode. The blue curve plots the normalized per-frame
  motion magnitude of the raw robot state, and the red staircase indicates the
  number of completed button presses from the subtask annotation. The model does
  not receive this hand-drawn curve; it receives the 30-step quantized state
  history and delta summary as text. The plot visualizes why this signal is
  useful when the same subtask phrase is repeated many times.}
  \label{fig:press_button_state_signal}
\end{figure}
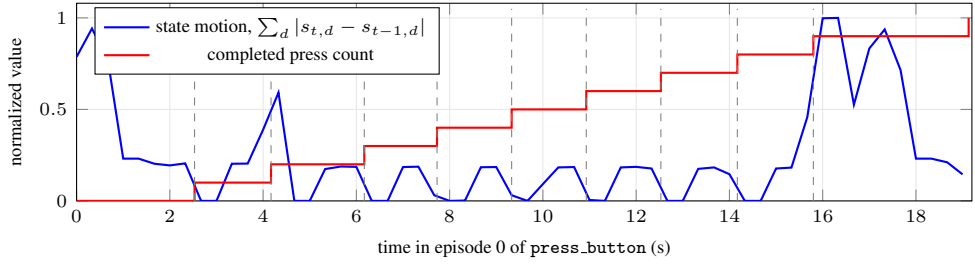

Figure~\ref{fig:press_button_state_signal} illustrates why this state channel is
important. In the RMBench \texttt{press\_button} task, the same action phrase
(``Press the middle button one time'') appears repeatedly, and the correct next
subtask depends on how many presses have already happened. Sparse visual
queries may only show the arm near a button, while the robot state contains
high-frequency joint and gripper motion at every control frame. We therefore do
not require System-2 to infer the count solely from isolated images; the
quantized state-history text exposes the recent actuation pattern.

\paragraph{Prompt variants and supervision target.}
Each training example is formatted as a chat-style instruction to Qwen-VL. The
input contains the global task goal, the input language memory, optional
additional guidance, the robot state history, and the multi-view observation
history. 

The assistant target is a JSON object:
\begin{equation}
  z_t =
  \left\{
  \texttt{"current\_subtask"}: a_{y(t)},\;
  \texttt{"active\_language\_memory"}: m_{y(t)}
  \right\}.
  \label{eq:json_target}
\end{equation}
The target memory is not a chain-of-thought trace; it is a concise semantic
state that can be fed back to System-2 at the next query.

Figure~\ref{fig:system2_prompt_template} shows the exact structure of the
training prompt. The example is taken from \texttt{press\_button}, where the
task requires pressing the left button twice, the middle button seven times,
and the confirm button once. For space, the state history is abbreviated; in
training, all 30 state rows and all selected state dimensions are emitted.

\begin{tcbverbatimwrite}{\jobname_sys2_prompt_template.txt}
SYSTEM:\\
You are a robot program for high-level manipulation. Given the global task goal,
an optional detailed global task instruction, an optional ordered subtask list,
the input language memory, and the history and current camera observations,
predict the subtask the robot should currently be in and the language memory
that should be active now. Return JSON only with keys "current\_subtask" and
"active\_language\_memory".\\
\par
USER:\\
Global Task Goal:\\
"Observe the two numbers on the table. Press the left button the number of times corresponding to the number on the left, and press the middle button the number of times corresponding to the number on the right. Then press the right button once to confirm."
\par
Input Language Memory:\\
I pressed the left button twice and the middle button six times.\\
\par
Robot State History\\
(causal, oldest to current, 30 state slots, stride=1 state step, 14-D, values
are 256-bin integers after tanh-squashing raw state to [-1,1]; -1 means
padded/missing state dimension):\\
t-29: 128 249 230  37 126 173 128 128 ...\\
t-28: 128 249 230  37 126 173 128 128 ...\\
...\\
t-2 : 128 249 230  37 126 173 128 128 ...\\
t-1 : 128 248 229  37 126 171 128 128 ...\\
t   : 128 247 228  37 126 169 128 128 ...\\
\par
Robot State Delta Summary:\\
per\_step\_l1\_x1000: 0 0 0 4 19 48 86 74 12 ...\\
changed\_dims\_per\_step: 0 0 0 2 4 5 6 5 2 ...\\
oldest\_to\_current\_delta\_x1000: 0 97 83 0 0 41 0 0 ...\\
state\_change\_run\_count: 1\\
current\_transition\_changed: 1\\
\par

...

\par
ASSISTANT:\\
\{"current\_subtask": "Press the confirm button.",\\
\phantom{\{} "active\_language\_memory": "I pressed the left button twice and the middle button seven times."\}
\end{tcbverbatimwrite}

\begin{center}
\begin{tcolorbox}[
    breakable,
    colback=gray!8,
    colframe=black!60,
    arc=0mm,
    boxrule=0.4pt,
    top=3mm,
    bottom=3mm,
    left=3mm,
    right=3mm
]
\fontsize{8.3pt}{10.4pt}\selectfont\ttfamily
\input{\jobname_sys2_prompt_template.txt}
\end{tcolorbox}
\captionof{figure}{Training prompt template for System-2. The prompt contains task-level
language, input memory, optional ordered subtask guidance, quantized causal
state history, delta summaries, and multi-view image tokens. The target is a
minimal JSON object rather than free-form reasoning.}
\label{fig:system2_prompt_template}
\end{center}

\subsection{Real-World Robot Experiments}
\label{app:real_robot}

\subsubsection{Hardware Setup}
\label{app:real_hardware}

Real-world experiments are conducted on an ARX ACONE dual-arm platform equipped with RGB cameras for scene observation. The robot is evaluated in tabletop manipulation scenes containing household objects, kitchen tools, and chemistry-style containers. Unless otherwise specified, all experiments are initialized from manually reset scene configurations, while subsequent subtask transitions are determined online by Cortex rather than by human intervention. System-1 runs closed-loop action generation at approximately 10 Hz, while System-2 performs high-level verification and subtask generation at approximately 2 Hz.

\subsubsection{System-1 Executor}
\label{app:real_sys1}

The real-world executor is instantiated as a MEM-style VLA policy \cite{torne2026mem}, denoted as $\pi_{\mathrm{mem}}^{\mathrm{sub}}$, with a short several-second memory window. It is trained on long-horizon demonstrations collected in the target robot setup, but its action prediction is conditioned on the current RGB observation, the short memory window, and the executable subtask accepted by the harness. In contrast to the task-level end-to-end baseline $\pi_{\mathrm{mem}}$, which is conditioned on the original long-horizon task instruction, System-1 receives a short, physically grounded command such as \textit{``pick up the glass beaker''} or \textit{``place the cup into the microwave''}. This conditioning suppresses irrelevant future steps and reduces ambiguity when multiple visually similar objects or repeated skill primitives appear in the same episode. Training hyperparameter are shown in the following table \ref{tab:realworld_pimem_training}.
  \begin{table}[h]
      \centering
      \small
      \caption{Real-world $\pi_{mem}$ executor training configuration.}
      \label{tab:realworld_pimem_training}
      \begin{tabular}{ll}
      \toprule
      \textbf{Item} & \textbf{Configuration} \\
      \midrule
      Model size & 3.62B parameters \\
      Backbone & PaliGemma, width 2048, depth 18 \\
      Action expert & width 1024, depth 18 \\
      Distributed training & DDP \\
      Precision & bfloat16 \\
      Optimizer & AdamW, $\beta=(0.9,0.95)$, $\epsilon=10^{-8}$ \\
      Learning rate & $5\times10^{-5}$ \\
      LR schedule & Cosine decay with min LR $5\times10^{-6}$ \\
      Warmup steps & 2,000 \\
      Weight decay & $1\times10^{-2}$ \\
      Global batch size & $4 \times 8 = 32$ \\
      Training step & 30,000\\
      Max text length & 200 tokens \\
      Action dimension & 32 \\
      Action horizon & 100 \\
      Temporal frames & 4 frames \\
      Temporal stride & 20 \\
      Temporal state tokens & Enabled \\
      Temporal attention & Every 4 layers \\
      \bottomrule
      \end{tabular}
  \end{table}
\paragraph{Executor training data.}
System-1 training data are segmented into subtask-level clips with paired language annotations and continuous action trajectories. For complex real-world tasks, manually calibrated boundaries are used for a subset of demonstrations to establish reliable supervision around contact-rich transitions such as grasping, pouring, placing, and tool release. The remaining trajectories can be aligned by the automated segmentation pipeline described in Section~\ref{app:dataset_details}. This hybrid strategy is intended to preserve execution precision while reducing the amount of manual annotation required for long-horizon data collection.

\subsubsection{Deployment Harness}
\label{app:real_harness}

Long-horizon execution requires close coordination between System-2 planning and System-1 control, while real-world deployment often introduces transient occlusions, delayed state changes, and unexpected perturbations. To stabilize this asynchronous dual-system loop, the deployment harness serves as a lightweight arbitration layer between the low-frequency planner and the high-frequency executor.

At each System-2 inference step, the harness receives a candidate subtask and memory update. Instead of forwarding every prediction directly to System-1, it performs command holding and abnormal-transition filtering. A new command is accepted only when it can be mapped to an executable System-1 command and does not exhibit abnormal high-frequency switching across adjacent System-2 predictions. Otherwise, the previous command is preserved, allowing System-1 to continue high-frequency action generation without discontinuous interruptions.

\paragraph{Standardized Command Mapping.}
Because System-2 generates open-ended language while System-1 is trained with standardized command annotations, the harness normalizes each generated subtask into an executor-compatible command. Exact template matching is used when the generated command follows a canonical skill template. For semantically equivalent but surface-level different expressions, such as \textit{``grasp the beaker''} and \textit{``pick up the beaker''}, keyword matching is used as a conservative fallback. When multiple executable labels are plausible, the harness selects the output according to the abnormal-switching filter described above. These standardization and conservative selection rules reduce distribution shifts caused by surface-form language variation, encouraging System-1 to condition on the intended subtask command rather than scene-specific biases in the training data.

\paragraph{Timeout-driven Action Processing.}
Real-world perception and execution are not perfectly synchronized. In some cases, System-1 may complete a motion before System-2 visually confirms the transition, causing the robot to pause near a boundary state. To prevent deadlock, the harness applies a timeout mechanism. If System-2 produces no accepted transition within a predefined interval, the robot executes a slow, low-amplitude corrective or reset motion. This motion does not force task progress; instead, it refreshes the visual evidence available to System-2 while minimally perturbing the task state. This mechanism is particularly important in the beaker washing and chemical liquid stirring tasks (Appendix~\ref{app:real_beaker_washing} and Appendix~\ref{app:real_stirring}), where the most informative head-view observation comes from a relatively high camera viewpoint and therefore differs from much of the training distribution. Under this harder zero-shot setting, the timeout-driven escape from local deadlocks substantially improves real-world reliability without requiring additional System-1 augmentation.

\subsubsection{Zero-Shot Deployment Capabilities}
\label{app:real_capabilities}

The following real-world experiments analyze two capabilities that are central to Cortex's zero-shot deployment: prompt-mode flexibility and subtask-level progress feedback. The former allows Cortex to accommodate different prompt specifications while preserving the correctness of the overall plan, whereas the latter stabilizes long-horizon execution by deciding when to preserve the current subtask, update memory, or advance the procedure.

\paragraph{Prompt-mode flexibility.}
Cortex supports coarse goals, detailed procedural instructions, and explicit subtask lists within a unified deployment interface. This flexibility is important in zero-shot settings because real-world tasks differ in how much procedural structure the prompt should provide. We use the trash disposal experiment as the primary case study for prompt-mode comparison (Fig.~\ref{fig:appendix_trash_prompt_modes}) because it is compact enough for controlled comparison while still requiring object grounding and completion verification.

\paragraph{Subtask-level feedback---enabling trial-and-error.}
Cortex maintains real-time feedback during physical execution. When the observed physical state deviates from the nominal plan, Cortex can identify the mismatch instead of blindly switching to the next subtask according to a fixed order or common procedural logic, or persistently maintaining the current subtask without reassessment. Benefiting from memory compression and temporal alignment during training, Cortex jointly considers the input memory, task instruction, and current observation to determine whether the current subtask has been completed and whether the state recorded in memory should be updated. It then updates both the executable subtask and the memory accordingly. This capability is particularly important under real-world perturbations, partial execution failures, or delayed visual verification.

\begin{figure}[htbp]
    \centering
    \includegraphics[width=1.0\textwidth]{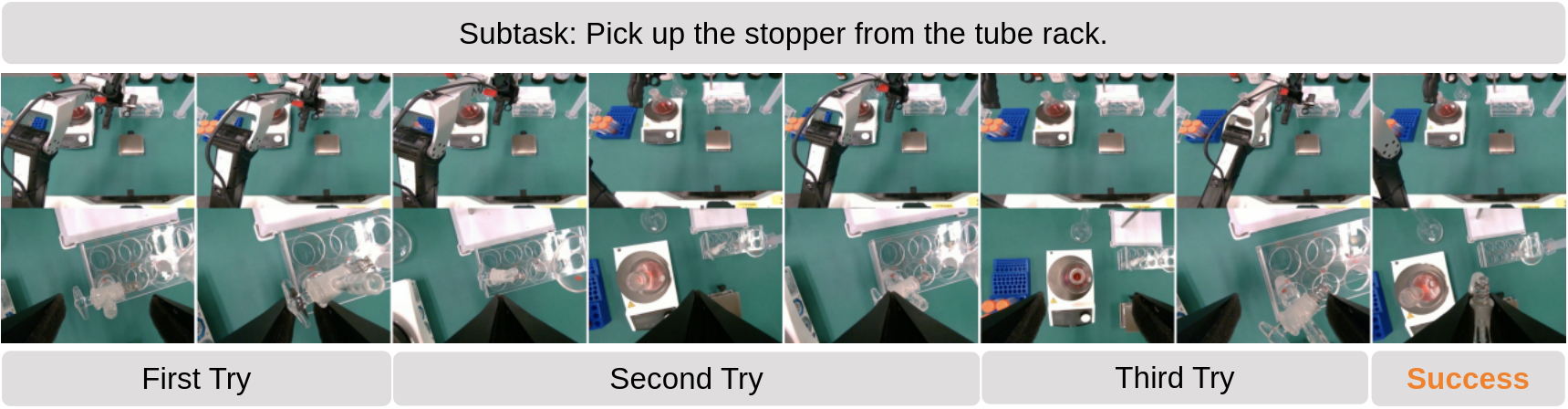}
    \caption{Local execution failure during stopper grasping. Cortex keeps the current subtask and memory until the grasp succeeds.}
    \label{fig:retry_stopper}
    \vspace{-5mm} 
    \end{figure}

We evaluate this capability through two representative cases: a human-induced state perturbation and a local execution failure during robot manipulation. In the perturbation case, the nominal task requires the robot to pick up a water bottle, open its cap, and pour water. Before execution, we manually open the bottle cap. Cortex observes this changed state and automatically appends the memory update \textit{``I opened the cap of the bottle''}. After picking up the bottle, it directly proceeds to the pouring subtask instead of redundantly outputting an open-cap command. In the local-failure case from the chemical liquid stirring task shown in Fig.~\ref{fig:retry_stopper}, the robot attempts to grasp a stopper from the tube rack. Because the stopper rotates under contact and becomes difficult to pinch, the first attempts fail. Cortex detects that the stopper has not been picked up, preserves the current memory and subtask, and only advances to the next subtask after the third attempt succeeds. These cases directly test whether Cortex performs online success verification rather than blindly following a precomputed subtask list.

% \paragraph{Composable subtask execution.}
% Cortex represents long-horizon behavior as a sequence of executable subtasks that can be recombined across tasks. The same primitive skills, such as picking, placing, moving to a target region, releasing, washing, or stirring, can be arranged into different procedures according to the global instruction and current memory. The beaker washing and chemical stirring experiments illustrate this behavior in longer and more constrained workflows. \textcolor{red}{@fdl: 这里如果不想显式声明“组合子任务执行”的独立实验，可以把这句改成更保守的能力描述，或者移到 limitation 里作为未来工作。}

\subsubsection{Experiment 1: Oven Heating}
\label{app:real_oven_heating}

The oven heating experiment represents a common kitchen task with only a few relevant objects and a clear causal order. The task requires the robot to open the oven door, pick up the dish, place it into the oven, and close the door. It mainly evaluates whether Cortex can switch between subtasks at the correct physical boundaries while preserving the logical dependency between consecutive actions.

\paragraph{Experiment setting.}\mbox{}\\[0.3em]
\textbf{\textit{Task instruction}}: ``Put the dish into the oven to heat it up.''

\begin{figure}[htbp]
\centering
\includegraphics[width=1.0\textwidth]{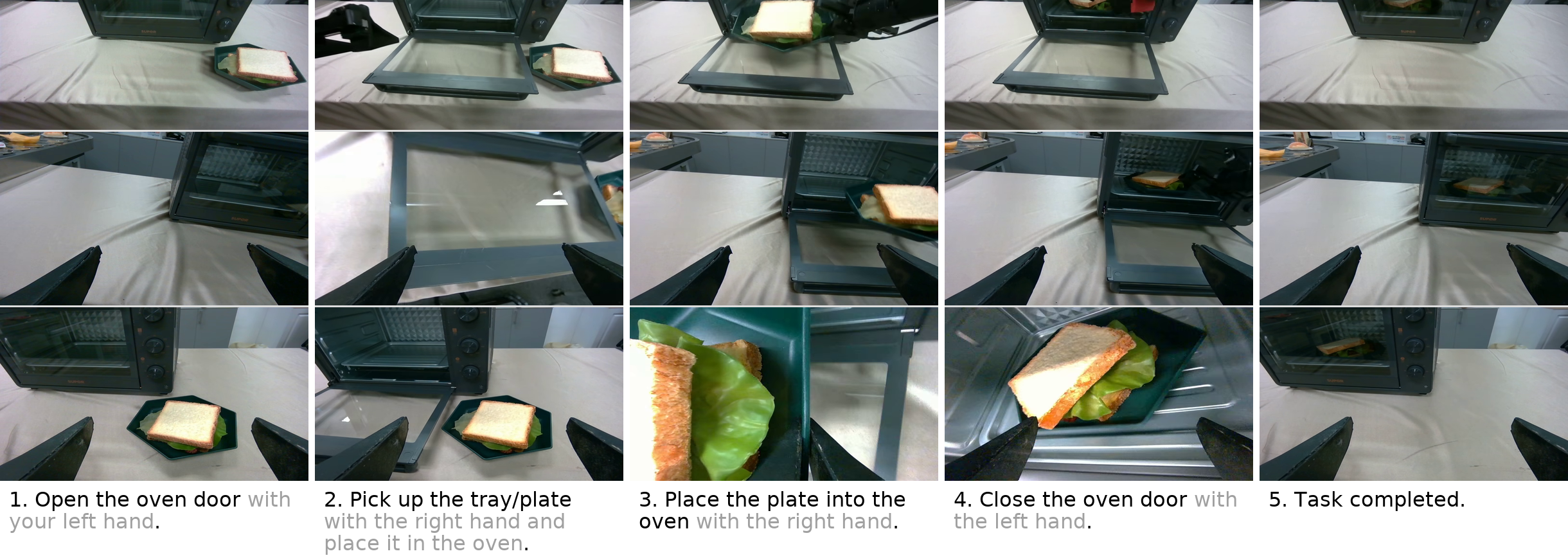}
\caption{Multi-view rollout of the oven heating experiment. From top to bottom, the rows show the head, left-wrist, and right-wrist views. Each column shows an action-switching frame, and the displayed output is the current subtask.}
\label{fig:appendix_oven_rollout}
\vspace{-5mm} 
\end{figure}

\paragraph{Qualitative results.}
For such common household tasks, Cortex can produce a correct task plan from a short input instruction and select appropriate transition times between subtasks. In general, the model switches to the next subtask only after confirming that the previous subtask has been fully completed. As shown in Fig.~\ref{fig:appendix_oven_rollout}, before switching to the second subtask, the oven door has already fallen completely onto the table surface rather than remaining in the air, indicating that System-2 waits for a stable completion state before issuing the command to pick up the plate.

\subsubsection{Experiment 2: Trash Disposal}
\label{app:real_trash}

The trash sorting experiment is an unordered manipulation task over multiple objects from similar categories. The core challenge is not procedural complexity, but whether the system can robustly identify each target object and generate expressions with a unique visual referent. We use this experiment to verify the effectiveness of the multi-mode System-2 interface and, based on these observations, further optimize the training of System-1.

\paragraph{Experiment setting.}\mbox{}\\[0.3em]
\textbf{\textit{Task instruction}}: ``Sort the garbage on the desktop into recyclable and non-recyclable.''

\textbf{\textit{Detailed task setting}}: ``Pick up the red plastic wrapper, the yellow can and the orange can with the right arm, place each item into the blue bin.''

\begin{figure}[htbp]
    \centering
    \includegraphics[width=1.0\textwidth]{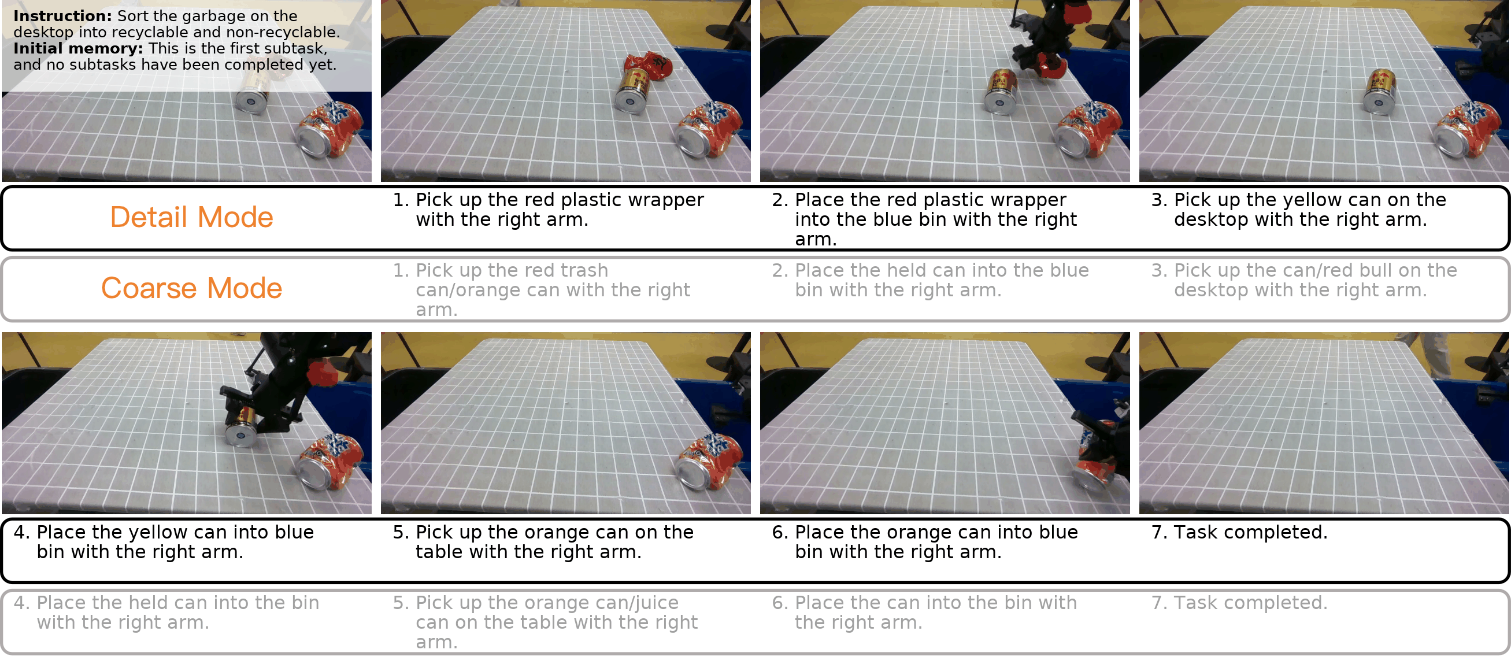}
    \caption{Prompt-mode comparison for trash disposal. Each frame is an action-switching frame, and the displayed output is the current subtask. The top row shows detailed mode, and the bottom row shows coarse mode.}
    \label{fig:appendix_trash_prompt_modes}
    \vspace{-5mm} 
    \end{figure}

\paragraph{Qualitative results.}
Figure~\ref{fig:appendix_trash_prompt_modes} shows that both prompt modes produce valid long-horizon executions, but they differ in how explicitly the intermediate subtasks are grounded. In the detailed mode, the generated commands are more object-specific and the action switches align closely with the intended pick-and-place sequence. In the coarse mode, the model must infer both the relevant targets and their execution order from the scene, leading to a more implicit decomposition of the same task.

\textbf{\textit{Prompt-mode qualitative comparison.}} This experiment compares prompt granularity under the same physical task. In detailed mode, the instruction explicitly names the red plastic wrapper, yellow can, orange can, and blue bin, so System-2 produces more object-specific subtasks and follows a cleaner pickup--place sequence. This mode is advantageous when the task requires precise target grounding, stable action ordering, and reliable completion checks across multiple similar objects. In coarse mode, System-2 receives only the high-level sorting goal and must infer the target objects and disposal order from the scene. This mode is more flexible and better reflects open-ended task understanding, because the system can decompose an underspecified goal into executable subtasks online. The comparison therefore shows a tradeoff: detailed mode improves precision and temporal stability, while coarse mode tests semantic generalization and autonomous task decomposition.

\subsubsection{Experiment 3: Beaker Washing}
\label{app:real_beaker_washing}

The beaker washing experiment evaluates a longer manipulation chain involving tool or container handling, placement near a washing area, and repeated state changes. Compared with trash disposal, this task places stronger requirements on memory because the beaker may need to be moved through multiple spatial regions before the final washed state is achieved.

\paragraph{Experiment setting.}\mbox{}\\[0.3em]
\textbf{\textit{Task instruction}}: ``Wash the beaker using the water from the bottle.''

\textbf{\textit{Detailed task setting}}: ``First, pick up the beaker, then place the beaker on the platform. Next, pick up the water bottle from the table. After opening the cap of the water bottle, place its cap on the table. Then, pour the water into the beaker and place the held water bottle back to the table. Later, pick up the beaker from the platform, pour water from the beaker into the kettle. Finally, place the beaker back on the platform.''

\begin{figure}[htbp]
    \centering
    \includegraphics[width=1.0\textwidth]{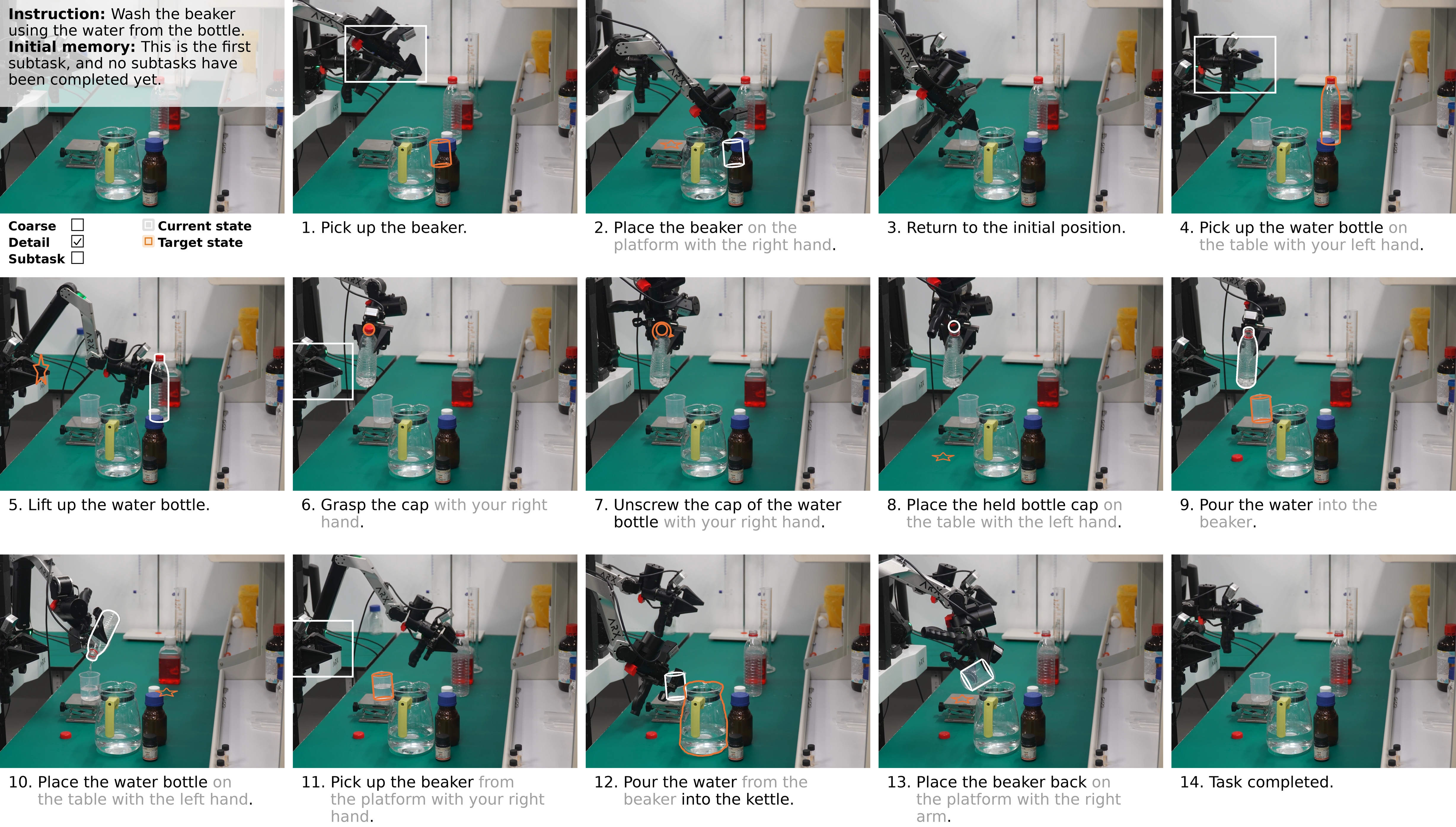}
    \caption{Detailed subtask prediction and execution process of Beaker Washing task}
    \vspace{-5mm} 
    \end{figure}
    
\paragraph{Qualitative results.}
The beaker washing task stresses both long-horizon state tracking and local execution stability. The robot must first move the beaker to the platform, manipulate the bottle and its cap, pour water into the beaker, return the bottle, and finally transfer the washed beaker state to the kettle-side operation. These stages contain repeated primitives such as picking, placing, and pouring, but the correct action depends on the accumulated procedural state rather than the current image alone.

Cortex handles this task by separating procedural tracking from local control. System-2 maintains the active memory over bottle state, cap state, beaker location, and washing progress, and only advances when the current subtask is visually verified. The executor $\pi_{\mathrm{mem}}^{\mathrm{sub}}$ therefore receives a localized command at each stage, which reduces ambiguity when the same object appears in multiple visually similar configurations.

\begin{figure}[htbp]
    \centering
    \includegraphics[width=1.0\textwidth]{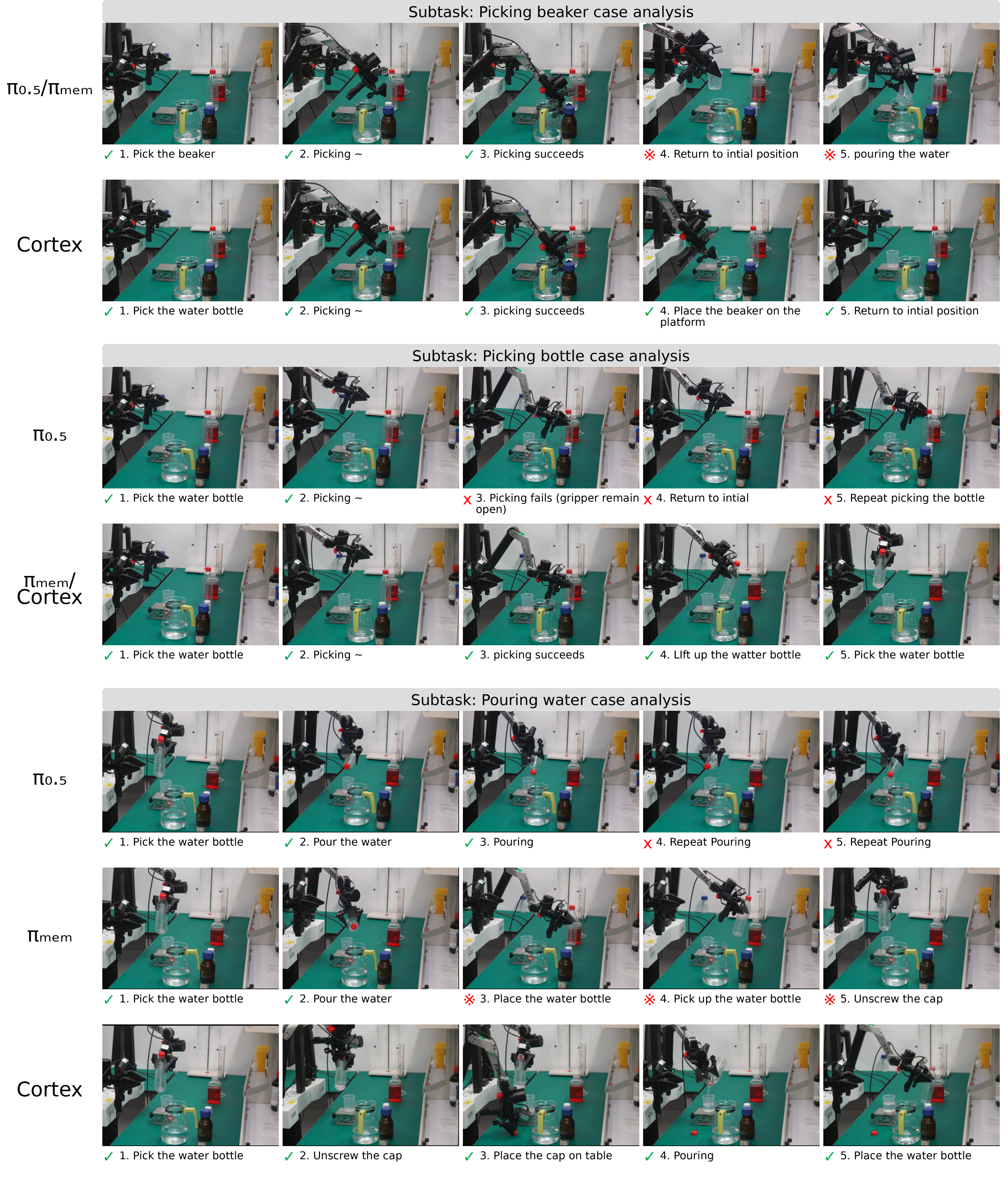}
    \caption{Performance comparison across different baselines in the beaker washing task. $\pi_{0.5}$ tends to repeat memory-sensitive reciprocal motions, while the task-level $\pi_{\mathrm{mem}}$ can confuse visually similar task phases; Cortex maintains the correct order through explicit subtask routing and transition verification. \textcolor{green}{\checkmark} indicates successful subtask execution, \textcolor{red}{$\times$} indicates failure and \textcolor{red}{※} indicates wrong subtask order.}
    \label{fig:app_beaker_e2e_comparison}
    \vspace{-5mm} 
    \end{figure}
    
\paragraph{Comparison with end-to-end VLA.}
Figure~\ref{fig:app_beaker_e2e_comparison} shows representative end-to-end failures in the beaker washing task. The first baseline, $\pi_{0.5}$, has no explicit memory. It can execute short primitives, but it often fails to decide when a reciprocal manipulation phase has ended. For example, after picking the bottle or entering a pouring-related motion, it may repeatedly revisit the same local behavior instead of committing to the next subtask, which prevents the full washing procedure from completing.

The second baseline, the task-level $\pi_{\mathrm{mem}}$, uses the same short several-second memory design as our executor and can therefore finish some short local sequences more reliably. However, conditioning it on the original long-horizon task instruction forces the policy to infer the current subtask stage from visual context alone. In this long-horizon task, several stages differ only subtly in the image, such as holding the bottle before versus after cap opening, or approaching the beaker before versus after water has been poured. As a result, $\pi_{\mathrm{mem}}$ may confuse the cap-opening, bottle-lifting, and pouring phases, causing out-of-order behavior such as attempting to pour before the cap has been opened.

In contrast, Cortex combines explicit subtask-memory tracking with $\pi_{\mathrm{mem}}^{\mathrm{sub}}$. System-2 preserves the intended procedure and dispatches only the currently executable subtask, while the low-level policy focuses on executing that localized command. This design keeps the task order stable even when adjacent stages are visually similar.

\subsubsection{Experiment 4: Chemical Liquid Stirring}
\label{app:real_stirring}

The liquid stirring experiment evaluates a chemistry-style workflow with rare objects, strict action order, and fine-grained success conditions. It is the most suitable task for detailed or subtask-mode prompts because the system must preserve procedural order while maintaining stable memory over multiple object interactions.

\paragraph{Experiment setting.}\mbox{}\\[0.3em]
\textbf{\textit{Task instruction}}: ``Pour liquid from the graduated cylinder into the round-bottom flask with the aid of a funnel, then place the flask on the stirrer and press the switch to start stirring.''

\textbf{\textit{Detailed task setting}}: ``First, pick up the funnel on the tube rack. Then insert it into the round-bottom flask on the platform. Next, pick up the graduated cylinder from the right side of the table and lift the held graduated cylinder close to the funnel. Then, pour the liquid from the graduated cylinder into the flask through the funnel. Place the held cylinder back on the table. Next, approach the top of the flask on the platform and split the upper part funnel. Place the held funnel back on the rack. Then, pick up the round-bottom flask from the table and place it on the black stirrer. Next, pick up the stopper from the tube rack. Insert the stopper into the round-bottom flask. Finally, press the white switch on the right bottom of the stirrer to turn it on.''

\paragraph{Qualitative results.}
The chemical liquid stirring task is more order-sensitive than the household tasks. It involves rare laboratory objects and a fixed procedure: inserting the funnel, lifting the graduated cylinder, pouring through the funnel, removing and returning the funnel, moving the flask to the stirrer, inserting the stopper, and pressing the switch. Many stages are contact-rich and visually subtle, so advancing too early can invalidate the remaining procedure.

Cortex maintains the current subtask during repeated or contact-rich motions and advances only after the relevant completion evidence is observed. This behavior is especially important for insertion and removal operations, where a small pose change can determine whether the object is actually inserted, lifted, or released. The memory state provides the long-horizon procedural context, while $\pi_{\mathrm{mem}}^{\mathrm{sub}}$ executes each verified local command.

\begin{figure}[htbp]
    \centering
    \includegraphics[width=1.0\textwidth]{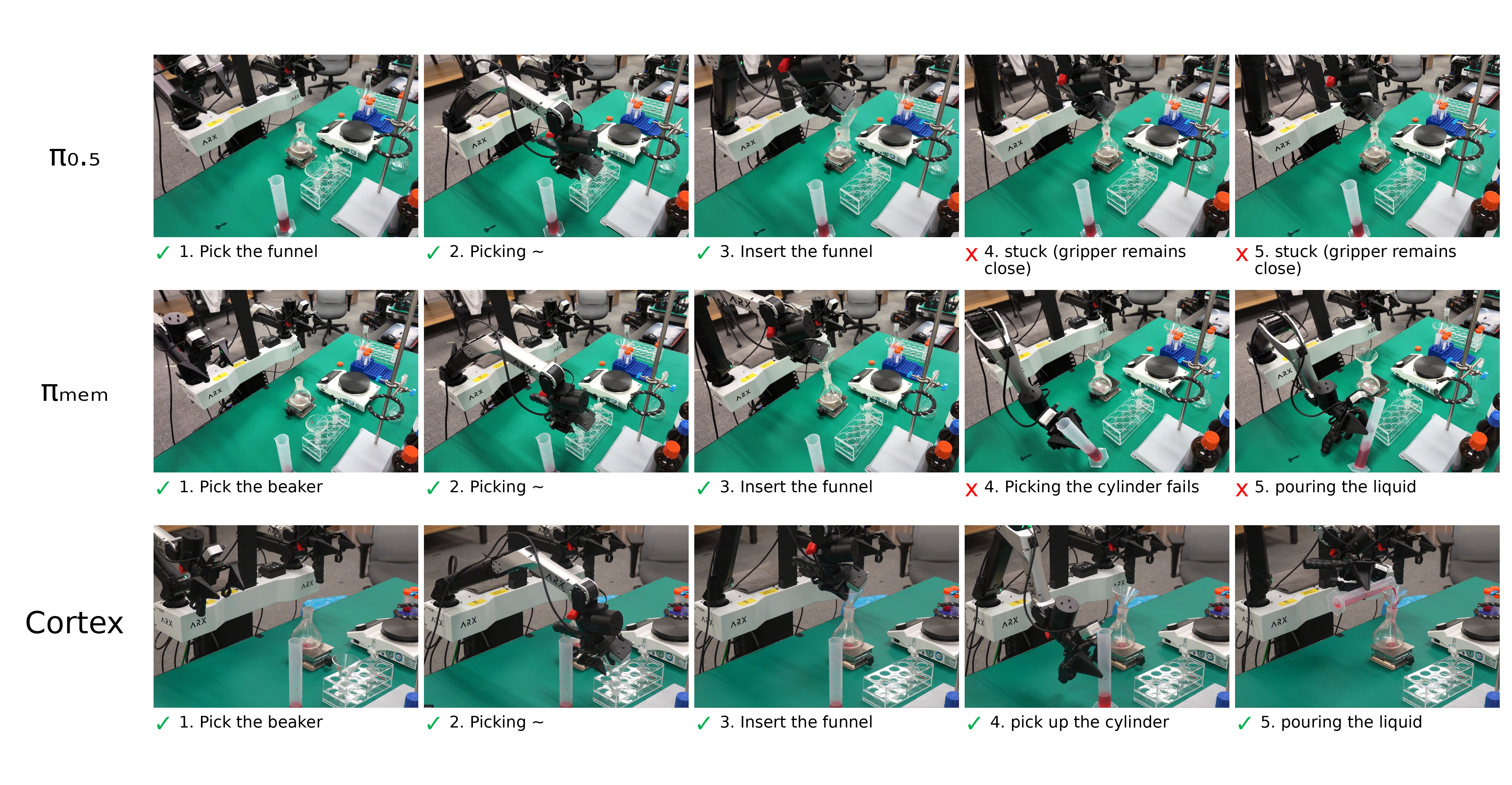}
    \caption{End-to-end failure cases in the chemical liquid stirring task. $\pi_{0.5}$ fails to terminate memory-sensitive insertion/removal phases, while the task-level $\pi_{\mathrm{mem}}$ can confuse adjacent procedural stages; Cortex preserves task order through explicit memory and subtask-conditioned execution. \textcolor{green}{\checkmark} indicates successful subtask execution, \textcolor{red}{$\times$} indicates failure and \textcolor{red}{※} indicates wrong subtask order.}
    \label{fig:app_chemical_e2e_comparison}
    \vspace{-5mm} 
    \end{figure}

\textbf{\textit{Retry behavior.}} When stopper grasping fails locally, Cortex preserves the active subtask and re-attempts execution until visual evidence supports completion, rather than prematurely advancing the procedure.
\paragraph{Comparison with end-to-end VLA.}
Figure~\ref{fig:app_chemical_e2e_comparison} summarizes representative chemical-task failures. Because $\pi_{0.5}$ does not maintain an explicit memory state, it struggles to terminate insertion and removal-style subtasks. In practice, it can remain stuck around local contact behaviors, such as repeatedly trying to insert or remove the funnel, because the current observation alone does not provide a reliable signal that the procedural stage has been completed.

The task-level $\pi_{\mathrm{mem}}$ reduces some of this short-term repetition, but it still conditions only on the the overall task instruction. This makes adjacent chemistry stages easy to entangle: the visual difference between different subtask may be small, while the required action order is strict. A typical failure is executing a pouring motion before the graduated cylinder has been properly lifted toward the funnel.

Cortex avoids these failures by using System-2 to maintain the long-horizon procedural state and by dispatching only the verified current subtask to $\pi_{\mathrm{mem}}^{\mathrm{sub}}$. This keeps the controller grounded to the correct stage while preserving the ability to repeat local motions until visual verification confirms completion.

\end{CJK*}
\end{document}